\documentclass{article} 
\usepackage{iclr2027_conference,times}
\usepackage{graphicx}

\usepackage{amsmath,amsfonts,bm}

\def\eqref#1{equation~\ref{#1}}

\def\1{\bm{1}}

\DeclareMathAlphabet{\mathsfit}{\encodingdefault}{\sfdefault}{m}{sl}
\SetMathAlphabet{\mathsfit}{bold}{\encodingdefault}{\sfdefault}{bx}{n}

\usepackage{booktabs}
\usepackage{hyperref}
\usepackage{url}
\usepackage{tikz}
\usetikzlibrary{arrows.meta,calc,patterns,shapes.geometric}
\usepackage{multirow}
\usepackage{colortbl}
\usepackage{array}
\definecolor{cflow}{HTML}{56B4E9} 
\definecolor{cspan}{HTML}{D55E00} 
\definecolor{ctraj}{HTML}{000000} 
\definecolor{cgap}{HTML}{999999}  
\newcommand{\lvec}[1]{\reflectbox{$\vec{\reflectbox{$#1$}}$}}
\title{Residual Motion Splatting for Continuous Space-Time Video Super-Resolution}
\iclrfinalcopy
\author{Vincent Herfeld\thanks{Corresponding author, \texttt{vincent.herfeld@enhancelab.fr}} \\
Enhance Lab \\
Univ. Grenoble Alpes, Inria, CNRS, Grenoble INP, LJK
\AND
Eva Coupeté \\
Enhance Lab \\
\AND
Julien Mairal \\
Univ. Grenoble Alpes, Inria, CNRS, Grenoble INP, LJK
\AND
Jean Ponce \\
Enhance Lab \\
École normale supérieure - PSL \\
}
\definecolor{lightgrayrow}{gray}{0.93}
\begin{document}
\maketitle
\begin{abstract}
Continuous space-time video super-resolution produces a high-resolution, high-frame-rate video from a low-resolution, low-frame-rate one at arbitrary spatial scales and temporal instants. Classical spatial multi-frame super-resolution has shown that detail beyond a single frame’s sampling limit can be recovered by aligning complementary observations with sub-pixel precision and accumulating them on a finer grid. Existing learning-based methods largely depart from this principle: some align and fuse frames only at low resolution before regressing a continuous high-resolution representation, while others interpolate each frame to the output resolution before splatting. In both cases, the per-frame samples are resampled before fusion, so their sub-pixel positions are lost before frames are combined. Restoring this principle in a learned continuous model raises two challenges: estimating sub-pixel motion at unobserved times, and handling holes and occlusions when splatting a coarse grid onto a finer one. We address the first by predicting per-pixel continuous trajectories that refine linear optical-flow paths with smooth residual motion, using features integrated over the full video sequence. We address the second by directly splatting consecutive low-resolution features in a learned latent space onto a high-resolution grid at their predicted sub-pixel positions with predicted visibility weights, before a decoder resolves remaining ambiguities. This makes temporal interpolation and multi-frame super-resolution a single operation. Our method improves both reconstruction fidelity and perceptual quality on multiple standard benchmarks over the state of the art, while running up to an order of magnitude faster.
\end{abstract}

\section{Introduction}

 Spatial frequencies above a single frame's Nyquist limit become recoverable when several observations of the same scene fall at distinct locations between integer pixel positions and accumulate on a finer grid. Classical multi-frame super-resolution~\citep{tsaihuang1984,iranipeleg1991,farsiu2004} is built on this mechanism, and video supplies motion that places every frame at a different spatial phase. 
 Unlike conventional methods~\citep{haris2020spacetime,zoomingslowmo,xu2021temporal}, which fix the spatial and temporal upscaling rates $s$ and $r$ once and for all at training time,  continuous space-time video super-resolution (C-STVSR)  algorithms learn a single model that can be used at inference time for any values of $s$ and $r$, including ones not seen during training.\\
 Recent C-STVSR models depart from the classical mechanism of splatting low-resolution pixels directly to their high-resolution sub-pixel positions. Implicit neural representation (INR) methods predict high-resolution representations of each frame \textit{before} splatting them to the interpolated time steps~\citep{videoinr,motif,bfstvsr}; recent sinusoidal representation based methods predict over spatio-temporal voxels a parametric representation of a high-resolution video with no explicit motion model~\citep{v-cube}; Gaussian splatting based methods rasterise primitives whose position, spread and colour are regressed from features fused under linear motion~\citep{gsstvsr}. None of them carries the original low-resolution samples to their own sub-pixel positions under an explicit motion model. We reinstate this mechanism, which SPMC~\citep{spmc} introduced for fixed-scale video super-resolution at observed frames, and extend it to unobserved instants: a single splat maps the encoder features onto their sub-pixel coordinates on the high-resolution grid, along continuous trajectories predicted per location.
 
Our method works in four stages: (1) a video encoder produces latent features that capture temporal context over multiple frames; (2) given off-the-shelf forward and backward optical flow estimates, cubic splines with learned parameters are used to predict sub-pixel residual motions at every low-resolution location in temporally interpolated frames; (3) visibility estimates are also predicted at each location to account for effects such as occlusion and disocclusion; (4) the learned features are splatted onto the high-resolution interpolated frames, and mapped back onto pixel values by a learned decoder.\\
In this formulation, restoration fidelity depends on trajectory and visibility accuracy. While classical spatial super-resolution pipelines estimate motion between discrete observed frames, continuous temporal models must predict motion at unobserved intermediate times. To model complex motion, we refine coarse linear flow estimates by predicting residual splines. Leveraging feature maps over a broad temporal context enables the network to adjust trajectories according to scene content. Unlike closed-form quadratic approaches~\citep{qvi, eqvi} that impose a uniform constant-acceleration assumption everywhere, our model can learn varying acceleration dynamics. Our contributions are:\smallskip \\
\noindent $\bullet$ We bring back the classical multi-frame super-resolution principle in a learned
    continuous model. Low-resolution features are splatted directly onto the high-resolution grid at
    their own sub-pixel positions, at any instant between input frames, and weighted by a visibility weight
    predicted together with their trajectory. Nothing is upsampled or warped at high resolution before
    fusion, so multi-frame super-resolution and temporal interpolation become a single operation.
    Existing C-STVSR methods instead fuse either before upsampling or after interpolating each frame
    to the output resolution. \smallskip \\
\noindent $\bullet$ Accumulating observations this way requires sub-pixel motion at instants that
    were never sampled. At each low-resolution coordinate, we predict a continuous cubic B-spline
    residual on pretrained optical flow, from features propagated across the input sequence. An
    inference-time experiment (Appendix~\ref{app:traj-frames}) shows that this trajectory benefits
    from context beyond the two frames around the queried instant: with four frames it recovers
    most of the full-sequence gain and exceeds a closed-form quadratic model given the same frames. \smallskip \\
\noindent $\bullet$ On GoPro and Adobe240, our method improves both fidelity and perceptual quality
    over the state of the art. It generalises to unseen space-time scales, is the best C-STVSR method
    on arbitrary-scale video super-resolution and in video frame interpolation, and is the most robust to unseen input noise. Because no stage runs at high resolution before the splat, it
    is 6--11$\times$ faster than the C-STVSR methods with public weights.
\begin{figure}
    \centering
    \includegraphics[width=\linewidth]{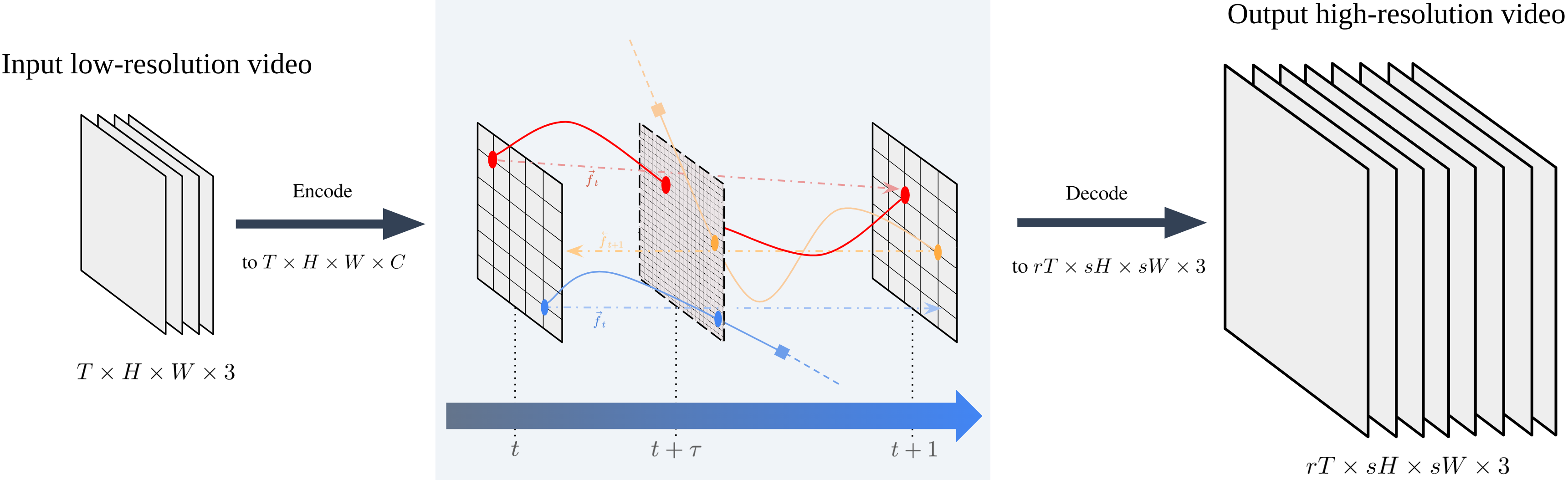}
    \caption{\textbf{Overview.} A video encoder turns the low-resolution input into features that
  carry context from the whole sequence. For each keyframe pair, every low-resolution feature is
  splatted onto the high-resolution grid at time $t+\tau$ (fine dashed grid) along a trajectory
  (solid) that corrects the linear optical-flow motion (dash-dotted) with a learned spline, and is
  weighted by a predicted visibility. Points visible in both keyframes (red) are fused from both, forward from $t$ and backward from $t+1$;
  points that disappear (blue) or appear (yellow) between the keyframes are taken from the
  frame in which they are visible, squares marking where visibility changes. A decoder maps the
  fused features to RGB frames.}
    \label{fig:enter-label}
\end{figure}
\section{Related Work}
\textbf{Classical Multi-Frame Super-Resolution.}
Super-resolution from several shifted
observations predates learned methods. Frequency-domain registration and
reconstruction~\citep{tsaihuang1984}, iterative back-projection~\citep{iranipeleg1991}
and robust regularised estimators~\citep{farsiu2004} all rest on the same premise:
the observations sample the scene at distinct sub-pixel offsets, and accumulating
them on a finer grid recovers aliased frequencies. Burst photography applies the same principle with kernel regression on a fine
grid~\citep{wronski2019handheld}, and early learned VSR kept it explicit: the SPMC
layer~\citep{spmc} maps low-resolution pixels onto the high-resolution grid at their
flow-displaced sub-pixel positions, performing motion compensation and upsampling in one step.
All of these estimate motion only between frames they actually observe, at a fixed output scale,
and degrade sharply where motion estimation fails, which is one reason later learned methods
replaced explicit accumulation with implicit feature fusion~\citep{basicvsrpp, rvrt}. In this work, we combine the strengths of both paradigms: we leverage implicit temporal feature extraction~\citep{rvrt} to construct rich context representations, which are then used to predict robust non-linear trajectories for explicit sub-pixel splatting at unobserved instants.\vspace{0.1cm}\\
\textbf{Fixed-Scale Space-Time Video Super-Resolution.}
Video Super-Resolution (VSR) and Video Frame Interpolation (VFI) traditionally
address spatial and temporal upsampling as independent tasks. VSR frameworks
aggregate information across frames with recurrent or transformer
backbones~\citep{edvr, basicvsr, basicvsrpp, rvrt, psrt, vsrm}, aligning them with
optical flow, deformable attention~\citep{deepdiveDCN, basicvsrpp} or
cross-attention~\citep{iart}. VFI methods synthesise intermediate frames along the
temporal axis using flow-based backward warping~\citep{superslomo, bao2019depth,
sim2021xvfiextremevideoframe} or kernel-based motion
compensation~\citep{niklaus2017video}. Unified fixed-scale Space-Time Video
Super-Resolution (STVSR) models~\citep{haris2020spacetime, zoomingslowmo,
xu2021temporal} fuse the two into single-stage architectures, avoiding the error
propagation and flicker of cascaded pipelines. All of them are bound to
pre-defined upsampling factors, through fixed pixel-shuffle heads~\citep{shi2016real}
or rigid temporal modulation blocks: the output scale and frame rate are decided
when the model is trained. \vspace{0.1cm}\\
\textbf{Arbitrary-Scale Spatial Super-Resolution.}
Continuous spatial scaling was pioneered through implicit neural
representations~\citep{liif, lte}, which parameterise images as continuous
functions over coordinate grids using MLPs or local Fourier
bases~\citep{pak2023b}. Continuous 2D Gaussian
Splatting~\citep{pixel2gaussian, gsasr} has since emerged as an efficient
alternative, rendering arbitrary spatial grids by forward rasterisation of fitted
primitives rather than by point-wise implicit coordinate queries. Extending
continuous evaluation to video, Arbitrary-Scale Video Super-Resolution (AVSR)
frameworks~\citep{liavsr, shangavsr} incorporate temporal guidance features, but
address spatial scaling only and cannot synthesise intermediate frames.\vspace{0.1cm}\\
\textbf{Continuous Space-Time Video Super-Resolution (C-STVSR).}
VideoINR~\citep{videoinr} scales both axes with decoupled spatial and temporal
implicit functions, but predicts a backward flow at unobserved times: that field
cannot be measured, and at occlusion boundaries it is unconstrained by the
observations. MoTIF~\citep{motif} replaces it with forward trajectories and
softmax splatting~\citep{softsplat}, regressing the displacement at time $t$ from
a SIREN; BF-STVSR~\citep{bfstvsr} keeps that pipeline and gives the time axis a
B-spline basis. Both learn the trajectory outright, unlike the closed-form fits
of the VFI lineage, which are cheap and stable on smooth motion but fixed in
order: linear~\citep{superslomo}, quadratic with an acceleration term recovered
from a third flow~\citep{qvi}, and least-squares-rectified quadratic~\citep{eqvi}. Both also evaluate the trajectory at the high-resolution
output coordinates, so the splat carries $N_\text{HR}$ feature tokens, and BF-STVSR
estimates it from two successive frames. A pair fixes a
displacement and says nothing about how it changes: acceleration is not
observable, so whatever curvature the spline expresses comes from the learned
prior rather than from the input. Other approaches have been explored such as using 3D Fourier Fields in V$^3$~\citep{v-cube} to map space-time signals
globally to bypass explicit motion, at the cost of detail under large
displacement which can exhibit chunk-boundary artefacts from the voxel partitioning their
memory footprint requires, or Gaussian splatting approaches such as GS-STVSR~\citep{gsstvsr} which regresses Gaussian
primitives from features already fused under linear motion rather than
transporting the observations. In our work, we instead predict the trajectory at each of the $N_\text{LR}$ low-resolution coordinates, from features propagated across the whole sequence, and as a residual on the flow estimate rather than as the full displacement.

\section{Method}
\label{sec:method}
Given a low-resolution image sequence $y \in \mathbb{R}^{T \times H \times W \times 3}$ with $T$ RGB frames of size $H \times W$, a
spatial upscale factor $s$ and a temporal upscale factor $r$, C-STVSR aims to
reconstruct high-resolution frames at continuous positions in space and time. 

Our method solves this problem in four stages.
We first estimate bidirectional optical flow across the input RGB sequence with an off-the-shelf estimator~\citep{raft}; motion in both temporal directions is needed to capture asymmetric dynamics and to resolve regions occluded in either the past or the future. These flows guide feature propagation in a video super-resolution backbone~\citep{rvrt}, which returns latent features $Y \in \mathbb{R}^{T \times H \times W \times C}$: every frame is encoded at its original spatial resolution but carries context from the whole sequence (Section~\ref{sec:encoder}). For each feature map $Y_t$, a network predicts (from its neighbouring feature maps $Y_{t+1}$ and $Y_{t-1}$), at every low-resolution position and in both temporal directions, two continuous functions of time: a B-spline correction to the flow-induced linear motion (Section~\ref{sec:trajectory}), and a visibility weight modelling whether the content is still visible at that instant or has been occluded. Both are predicted once per input frame, independently of the output frame rate, and are then
  evaluated at the intermediate times $t+\tau$ and $t-1+\tau$; to produce $r$ frames per interval
  we take $\tau \in \{\tfrac{j}{r}\}_{j=0}^{r-1}$. A splat then places the encoded features onto a
  fixed $4H \times 4W$ grid at the positions their corrected trajectories reach at time $\tau$,
  weighted by the predicted visibility (Section~\ref{sec:splatting}), so that temporal
  interpolation and multi-frame super-resolution happen in a single operation. A decoder finally maps the
  fused features to pixels, and for $s \neq 4$ the RGB frame is bicubically interpolated up or down to reach the factor $s$ (Section~\ref{sec:decoding}).
  Two properties characterise this design. First, every quantity is estimated in the low-resolution
  source domain, so the trajectory and visibility networks run at the input resolution. Second, the
  output is continuous in both dimensions: time enters through the spline evaluation at any
  $\tau$ in $[0,1]$ and space through the final resampling, so one trained model serves arbitrary
  space-time upsampling.

\subsection{Video-Aware Feature Extraction}
\label{sec:encoder}
Estimating interframe trajectories from only two adjacent temporal frames restricts motion modelling to constant velocity. Without access to broader temporal context, the network cannot observe higher-order motion dynamics such as acceleration or curvature; as a result, any non-linear trajectory it predicts relies on learned inductive biases rather than observed evidence. While parameterizations such as the B-spline motion model in BF-STVSR~\citep{bfstvsr} provide an effective continuous trajectory formulation, estimating its parameters from adjacent two-frame windows remains a key bottleneck. We retain the continuous B-spline parameterization but estimate it from features that aggregate long-range temporal context and use it to correct flow-based linear motion estimates. Enlarging the temporal context for the trajectory estimator improves reconstruction quality, indicating that the additional frames provide usable motion evidence (Appendix~\ref{app:traj-frames}). We adopt RVRT~\citep{rvrt}, a recurrent video super-resolution backbone, to propagate features bidirectionally across the entire sequence using optical flow maps estimated with RAFT~\citep{raft}. We replace RVRT's original fixed-scale upsampling block with our proposed residual motion multi-frame splatting.

Throughout the rest of the paper, right and left arrows over a symbol denote quantities associated with the forward and backward temporal directions, respectively; for example $\vec{f}_t$ and $\lvec{f}_t$ for the flows from the frame at time $t$ to the frame at times $t+1$ (forward) and $t-1$ (backward) respectively. 

\subsection{Trajectory and Visibility Estimation from Propagated Features}\label{sec:trajectory}
\begin{figure}[t]
\centering
\begin{tikzpicture}[
    scale=0.85,
    >=Latex,
    every node/.style={font=\small},
    dot/.style={circle, draw=black, fill=white, line width=0.9pt, inner sep=1.7pt},
]

\useasboundingbox (-1.2, -1.4) rectangle (12.2, 2.1); 

\begin{scope}[shift={(-1.0,0)}]
\coordinate (ut)     at (0,0);
\coordinate (utau)   at (2.5,0.95);
\coordinate (ut1)    at (6.3,-0.4);
\coordinate (uflow1) at (6.1,-1.05);

\coordinate (plin) at ($(ut)!0.45!(uflow1)$) ;

\draw[ctraj, line width=1.1pt, ->]
    (ut) .. controls (0.9,1.15) and (1.8,1.35) .. (utau)
         .. controls (3.3,0.5)  and (4,-2.2) .. (ut1);

\draw[cflow, line width=1.1pt, densely dashed, ->] (ut) -- (uflow1);

\draw[cspan, line width=1.3pt, -Stealth] (plin) -- (utau);
\node[circle, fill=cflow, draw=none, inner sep=1.1pt, label={below:\color{cflow}{$\vec{u}_{t+\tau}^{\text{flow}}$}}] at (plin) {};

\draw[cspan, line width=0.9pt, -Stealth] (uflow1) -- (ut1);
\node[circle, fill=cflow, draw=none, inner sep=1.3pt, label={below:\color{cflow}{$\vec{u}_{t+1}^{\text{flow}}$}}] at (uflow1) {};

\node[dot, label={above left:$\vec{u}_t$}]         at (ut)   {};
\node[dot, label={above:$\vec{u}_{t+\tau}$}]      at (utau) {};
\node[dot, label={right:$\vec{u}_{t+1}$}]          at (ut1)  {};

\node[cspan, anchor=east] at ($(plin)!0.5!(utau) + (-0.1,0.05)$) {$\vec{s}(\tau, u_t)$};
\node[cspan, anchor=west] at ($(uflow1)!0.4!(ut1) $) {$\vec{s}(1, u_t)$};
\node[cflow] at (1.2,-0.5) {$\vec{f}_t$};

\node[black!60, font=\normalsize] at (2.5, 1.8) {(a) motion trajectory}; 

\end{scope}

\draw[black!40, line width=0.8pt] (7.1,-1.2) -- (7.1,1.8);

\node[anchor=north west, inner sep=0] (splatimg) at (7.6, 1.4) {\includegraphics[height=2.6cm]{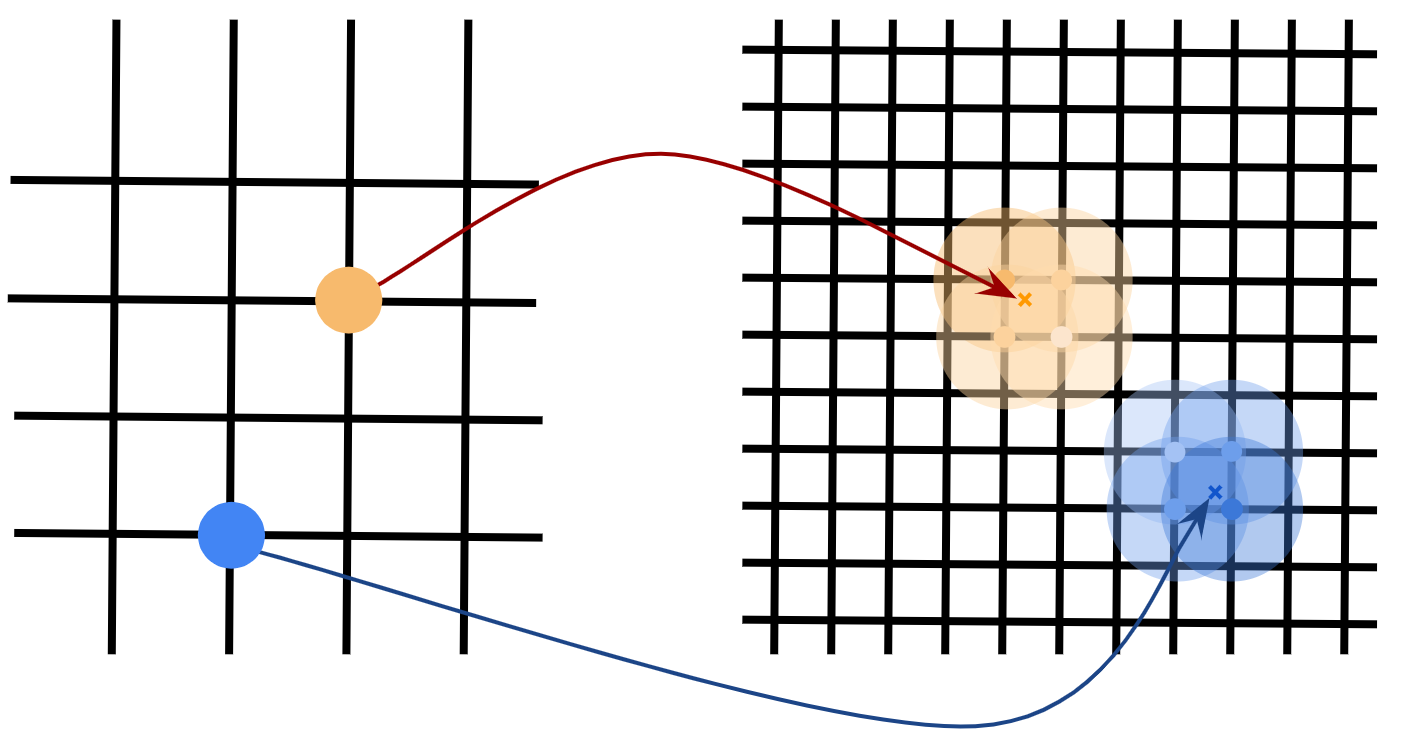}};

\node[black!60, font=\normalsize, anchor=center] at ($(splatimg.north)+(0, 0.4)$) {(b) sub-pixel splatting};

\end{tikzpicture}
\caption{(a) The full forward trajectory decomposed as the linear motion estimate from the optical-flow $\vec{f}_t$ corrected by a B-spline residual $\vec{s}$. (b) An illustration of the sub-pixel splat-and-upscale operation. We splat the content of a low-resolution pixel (left-grid) to a sub-pixel position in the high-resolution image (right-grid). Each of the four neighbours of the landing position receives a contribution that depends on its distance to the landing position, and is then spread by a Gaussian kernel.}
\label{fig:motion-curve}
\end{figure}

In the forward direction, a point located at position $u_t$ in the feature map
$Y_t$ is mapped by our forward model onto the point
\begin{equation}
    \vec{u}_{t+\tau} = u_t + \tau\,\vec{f}_t(u_t) + \vec{s}\big(\tau, \lambda, \vec{\mu}_t(u_t)\big),
\end{equation}
at time $t+\tau$ with $\tau$ in $[0,1]$, where $\vec{s}$ is the forward residual
B-spline (we write $\lvec{s}$ for the backward residual B-spline), parameterised by $\tau$, its bandwidth $\lambda$ taken constant over all
pixel locations and time intervals and learned with the other model parameters,
and the coefficients $\vec{\mu}_t(u_t)$ combine the two-dimensional control points
and the scalar knots associated with the spline, and depend on time $t$ and pixel
location $u_t$. To make sure the trajectories start at the correct locations, we design the splines to not contribute at the origins ($\tau =0$ for forward and $\tau =1$ for backward). The control points and knots are computed for all locations in a
single forward pass of the network $g_\theta$, as an $H \times W \times K \times 3$
tensor
\begin{equation}
    \vec{\mu}_t = g_\theta(Y_t, Y_{t+1}, \vec{f}_t).
\end{equation}
Although $g_\theta$ operates on the pair $(Y_t, Y_{t+1})$, both feature maps carry long-range sequence context. Unlike methods with limited temporal support that rely on rigid assumptions~\citep{qvi} or learned motion priors~\citep{motif, bfstvsr}, our splines predict plausible trajectories from observed multi-frame dynamics (Appendix~\ref{app:traj-frames}). By formulating these splines as a residual correction to linear optical flow, the network focuses on modelling non-linear motion and on correcting flow-estimation errors (at $\tau=1$), as illustrated in Figure~\ref{fig:motion-curve}(a)..

Symmetrically, the backward trajectory is $
    \lvec{u}_{t - 1  + \tau} =  u_t + (1-\tau)\,\lvec{f}_t(u_t) + \lvec{s}\big(\tau, \lambda, \lvec{\mu}_t(u_t)\big),$
where the backward coefficients are estimated from the backward feature context
and flow field by the same network $
    \lvec{\mu}_t = g_\theta(Y_t, Y_{t-1}, \lvec{f}_t).$

These positions define where the feature point $Y_t(u_t)$ will be splatted in the time-interpolated high-resolution feature maps $X_{t+ \tau}$ and $X_{t - 1 + \tau}$.

Splatting features from several source frames onto a single target grid is a
many-to-one map: whenever two contradicting sources, such as an occluder and the surface behind it, land on the same target pixel their contributions are accumulated and the result is a ghost. Resolving this requires a weight per low-resolution input feature deciding which should contribute at every moment in time (Appendix~\ref{app:visibility}). Following BF-STVSR~\citep{bfstvsr}, which predicts its splatting importance jointly with the trajectory, we read the weights from the trajectory as a learned linear projection of the B-spline basis with a final sigmoid layer $
    \vec{w}_t(u_t, \tau) = P_\varphi(\tau, \lambda, \vec{\mu}_t(u_t)) \text{ and } \lvec{w}_t(u_t, \tau) = P_\varphi(\tau, \lambda, \lvec{\mu_t}(u_t))$. These are $H \times W \times 1$ tensors with values in $(0,1)$ and weight the features during splatting.

\subsection{Splatting video features onto the high-resolution grid}
\label{sec:splatting}
Once the forward and backward trajectories are estimated, we use them to place the
  low-resolution features $Y_t$ and $Y_{t+1}$ directly onto a fixed high-resolution grid of scale
  factor $4$, producing the fused feature map $X_{t+\tau}$ that the decoder turns into pixels. The
  input features live on the low-resolution grid of $N_{LR} = HW$ points with normalised
  coordinates $\{u_n\}_{n=1}^{N_{LR}} = \{(\tfrac{i}{H},\tfrac{j}{W}) : i \in \{0,\dots,H{-}1\},\
  j \in \{0,\dots,W{-}1\}\}$, and the target features on the high-resolution grid of
  $N_{HR} = 16HW$ points with coordinates $\{v_n\}_{n=1}^{N_{HR}} = \{(\tfrac{i}{4H},
  \tfrac{j}{4W}) : i \in \{0,\dots,4H{-}1\},\ j \in \{0,\dots,4W{-}1\}\}$. We splat to a single $\times4$ grid, the scale the model is trained at, and resample the
decoded result to reach the requested scale (Section~\ref{sec:decoding}). Three reasons motivate
this choice. First, splatting to a larger grid brings little: each output pixel then receives
fewer observations, especially with only two source frames, and the gain of multi-frame
reconstruction diminishes as the scale factor grows
\citep{baker2002, lin2004, wronski2019handheld}. Second, for scales below $\times4$, the finer
grid places observations at least as precisely as needed, and filtering it down to the target
grid causes no loss of accuracy in our experiments (Table~\ref{tab:reds-avsr}). Third, a single
scale lets the decoder learn one sampling density rather than several.
  
In the following, we describe only the forward direction; the backward direction is symmetric. Each trajectory is a set of points $\{\vec{u}_{n,t+\tau}, \; \forall \tau \in [0,1]\}$ that depends on the source frame $Y_t$ and the starting low-resolution position $u_n$. Each point of the trajectory generally falls between four neighbouring high-resolution discrete points rather than on one of them; this is the sub-pixel landing position the trajectory is designed to predict. To respect this sub-pixel placement, we distribute the source feature $Y_t(u_n)$ over these four neighbours using bilinear weights, so each neighbour's share depends on its distance to the predicted landing point. Due to the small number of source pixels compared to the number of possible target positions, bilinear weights alone leave holes between splatted samples. We therefore widen each sample's footprint by composing the bilinear kernel with a Gaussian of standard deviation $\sigma_s = 1$ high-resolution pixel, so that target pixels between samples still receive a contribution.
Figure~\ref{fig:motion-curve}(b) illustrates the coverage of the resulting kernel $\kappa$.

The fused feature at every high-resolution position $v$ is the sum, over both source frames $Y_t$ and $Y_{t+1}$, of every source point's feature weighted by the visibility maps $\vec{w}$ and $\lvec{w}$ and by the kernel $\kappa$ evaluated at the distance to its predicted landing position:
\begin{equation}
X_{t+\tau}(v) = \sum_{n=1}^{N_{LR}} \vec{w}(u_n, \tau) Y_t(u_n) \kappa\big(v - \vec{u}_{n, t + \tau}\big) + \sum_{n=1}^{N_{LR}} \lvec{w}(u_n, \tau) Y_{t+1}(u_n) \kappa\big(v - \lvec{u}_{n, t + \tau}\big)  .  
\end{equation}
The first term accumulates features splatted from frame $t$ along its forward trajectory, the second from frame $t{+}1$ along its backward trajectory. For simplicity, we omit the normalization by the total weight landing at $v$, which makes $X_{t+\tau}(v)$ a weighted average of all contributions. This operation builds interpolated frames at continuous times and upscales them in a single pass, adding detail restoration beyond what the video encoder provides. It follows the same principle as classical multi-frame super-resolution~\citep{tsaihuang1984}, now in an end-to-end learned latent feature space evaluated at any continuous time $t + \tau$.

\subsection{Decoding}
\label{sec:decoding}
A decoder $D_\phi$ maps the splatted feature map $X_{t+\tau}$ to the high-resolution RGB frame: 
\begin{equation}
    x_{t+\tau} = D_\phi(X_{t+\tau}) \in \mathbb{R}^{4H \times 4W \times 3}.
\end{equation}
Splatting and decoding operate at a fixed factor of $4$ (Section~\ref{sec:splatting}); any target
scale $s$ is reached by bicubic up or downsampling of $x_{t+\tau}$ in RGB space, so observations are
accumulated at their sub-pixel positions before any interpolation.

\subsection{Training objective}
Let $\mathcal{D} = \{(y^b, x^b)\}_{b=1}^{B}$ denote a batch of $B$ video sequences, where $y^b$ represents the low-resolution input video and $x^b$ denotes the corresponding ground-truth high-resolution video. Each video $x^b$ consists of $N = (r(T-1)+1)s^2HW $ pixel values with $T \times H \times W \times 3$ the shape of $y^b$ and $r$ the temporal scale factor and $s$ the spatial scale factor (we train at $s=4$ and $r=8$).

We evaluate the reconstruction error using the Charbonnier distance~\citep{charb}:
\begin{align}
\forall (z_1,z_2) \in \mathbb{R}^2, \; \ell(z_1,z_2) = \sqrt{(z_1 - z_2)^2 + 10^{-6}},
\end{align}
such that the empirical reconstruction risk $R_{\text{rec}}(\Theta)$ over the batch is defined as:
\begin{equation}
R_{\text{rec}}(\Theta) = \frac{1}{BN} \sum_{b=1}^{B} \sum_{n=1}^{N} \ell\!\left(x^b[n], f(y^b; \Theta)[n]\right)
,\end{equation}
where $f(y^b; \Theta)$ denotes the video reconstructed by the network with weights $\Theta$.

\section{Experiments}
\subsection{Experimental setup}
\label{sec:setup}

\textbf{Implementation.}
We follow~\citet{v-cube} and use their hyper-parameter choices for the RVRT~\citep{rvrt} video encoder, with the embedding dimension set to 90 and 12 attention heads, and train it from scratch end-to-end. The encoded features have 64 channels. For the trajectory estimator $g_\theta$ we use a three-layer SIREN~\citep{inrper} with layer dimensions set to 64,64, 256. Following the choice of~\citet{bfstvsr}, we choose $K=64$ for the size of the B-spline basis. We use a Residual U-Net~\citep{resunet} to parameterise the decoder network $D_\phi$. \vspace{0.1cm}\\
\textbf{Training Details.}
We train the model end-to-end and optimise the model parameters with AdamW ($\beta_1=0.9,\beta_2 =0.99$) and use a cosine annealing scheduler over 500k iterations with an initial learning rate of
$10^{-4}$. 
A training sample is an input sequence of $14$ frames at $64\times64$ with a fixed temporal scale factor of $r=8$ and spatial scale factor of $s=4$. We choose a batch size of 3 and train on a single RTX PRO 6000 Blackwell GPU. The pretrained RAFT flow estimator is kept frozen throughout the first $90\%$ of the iterations and then fine-tuned with a ten-times-smaller learning rate. This first lets the trajectory model train stably and then allows the optical-flow model to specialize for the task in an end-to-end learning setting.\vspace{0.1cm}\\
\textbf{Datasets.}
 We train our model using the Adobe240~\citep{adobe240} dataset, a set of 133 RGB videos captured at 240 frames per second. Following the split from~\citet{videoinr} that is used in the works that followed~\citep{motif, bfstvsr, gsstvsr, v-cube}, we have 100 videos for training, 16 for validation and 17 for testing. The dataset contains various scenes and camera motion, and is the standard training set for
continuous space-time video super-resolution.
Low-frame-rate inputs are synthesised by keeping one frame in eight such so that consecutive input frames contain 30 fps motion. We also use Vid4~\citep{viddataset} and GoPro~\citep{gopro} datasets as test sets for the spatio-temporal task and the REDS~\citep{reds} dataset for the spatial-only task. 

\subsection{C-STVSR Performance}
\label{sec:cstvsr-results}
We first evaluate in the training setting ($\times4$ space, $\times8$ time), then on unseen scales.
Spatial SR is better posed than temporal SR: motion provides sub-pixel-shifted views of the same content,
whereas intermediate instants are never observed. A model can only predict a plausible trajectory, and
pixel metrics penalise a sharp but offset frame more than a blurred one. We therefore report PSNR computed over luminance (PSNR-Y)
alongside LPIPS~\citep{lpips} (SSIM over luminance is reported in Appendix~\ref{app:additional-metrics}).  FPS is measured on full GoPro
frames (Appendix~\ref{app:runtime}).\vspace{0.1cm}\\
Our method is best on GoPro and Adobe240 datasets on both subsets and metrics (Table~\ref{tab:sf4tsf8-psnry-lpips}),
improving on V$^3$ by up to $0.6$\,dB and on V$^3$-Large by up to $0.5$\,dB with 20\% fewer parameters.
The gain is largest in LPIPS: 12--17\% lower than V$^3$, whereas V$^3$'s ${\approx}2$\,dB gain over BF-STVSR
brought only 3--10\%. Replacing the trajectory at inference changes PSNR by up to $1.5$\,dB but LPIPS by
under $0.006$ (Appendix~\ref{app:traj-frames}): the trajectory controls where content lands, while our reconstruction stage (sub-pixel splatting
and the HR decoder) controls how sharply it is rendered. 

Trajectory errors therefore cost PSNR, not sharpness;
in one clip our sharper crops trail V$^3$ in PSNR until a 4--6 pixel shift is compensated
(Appendix~\ref{app:qualitative}). V$^3$ shares our RVRT encoder and RAFT fine-tuning, and trains with far more compute (2.5M iterations, batch 16,
16 GPUs vs.\ 500k, batch 3, 1 GPU), which points to the
reconstruction stage as the source of the gap. On Vid4, whose inputs are twice as far apart as in training, our method is on par with V$^3$. We are an order of magnitude faster than V$^3$ and about $6\times$
faster than MoTIF and BF-STVSR.

\textbf{Extreme scale factors.}
Following~\citet{motif, bfstvsr}, we test combinations of spatial and temporal scales on GoPro,
and add a two-step baseline that runs EMA-VFI~\citep{ema-vfi} followed by LIIF~\citep{liif}.
Table~\ref{tab:varying-scales} shows that our method is best for every combination, including
temporal scales up to twice and spatial scales up to three times those used in training. We evaluate V$^3$ with the authors' official inference code and released weights.
\begin{table}[th]
\caption{C-STVSR at $\times4$ spatial scale, $\times8$ temporal (GoPro, Adobe240) and $\times2$
temporal (Vid4). We report PSNR-Y$\uparrow$ / LPIPS$\downarrow$. \emph{Center} evaluates
frames 1, 5, 9 of each 9-frame sequence, \emph{Average} all output frames. Methods marked $^\dagger$ have no public weights, so we report their published PSNR-Y only. We highlight the best result in \textbf{bold} and the second best with \underline{underline}.}
\label{tab:sf4tsf8-psnry-lpips}
\centering
\resizebox{\textwidth}{!}{%
\begin{tabular}{l | c | c | c c c c c}
\hline
 Method & \# Par. & FPS & Vid4 & \parbox{2cm}{\centering {GoPro}\\\it Center} & \parbox{2cm}{\centering {GoPro}\\\it Average} & \parbox{2cm}{\centering {Adobe}\\\it Center} & \parbox{2cm}{\centering {Adobe}\\\it Average} \\
\hline
MoTIF & 12.6M & \underline{2.5} & 25.76 / 0.3024  &31.04 /  0.1554 & 30.04 /  0.1587 & 30.63 / 0.1408 & 29.82 / 0.1457 \\
BF-STVSR & 13.5M & 2.2 & 25.84 / 0.2931 &31.17 /  0.1507 & 30.22 / 0.1543 & 30.83 /  0.1313 & 30.12 /  0.1350  \\
GS-STVSR$^\dagger$ &12.7M&  \phantom{0.}\text{---}\phantom{00} & 26.04 / \phantom{0.}\text{---}\phantom{00} &31.33 / \phantom{0.}\text{---}\phantom{00}& 30.35 / \phantom{0.}\text{---}\phantom{00} & 31.13 / \phantom{0.}\text{---}\phantom{00} & 30.35 / \phantom{0.}\text{---}\phantom{00} \\
V$^3$ & 13.7M & 1.3 & 26.76 / \textbf{0.2649} & 32.96 / 0.1458 & 32.26 / 0.1485 & 32.91 / 0.1186 & 32.29 / 0.1223\\
V$^3$- Large$^\dagger$ & 20.6M & \phantom{0.}\text{---}\phantom{00} & \underline{26.82} / \phantom{0.}\text{---}\phantom{00}&  \underline{33.09} / \phantom{0.}\text{---}\phantom{00} & \underline{32.36} / \phantom{0.}\text{---}\phantom{00} & \underline{33.08} / \phantom{0.}\text{---}\phantom{00} & \underline{32.45} / \phantom{0.}\text{---}\phantom{00} \\
\rowcolor{lightgrayrow} 
\textbf{Ours} & 16.5M & \textbf{14.5} & \textbf{26.87} / \underline{0.2760}  & \textbf{33.40} /  \textbf{0.1261} & \textbf{32.54} / \textbf{0.1306} & \textbf{33.55} / \textbf{0.0986} & \textbf{32.78} / \textbf{0.1032}\\ 
\hline
\end{tabular}%
}
\end{table}
\begin{table}[htbp]
\centering
\caption{Performance comparison on the C-STVSR baselines for extreme scale factors on GoPro dataset. Results are evaluated using PSNR-Y (dB) and SSIM-Y metrics. All frames are interpolated by a scaling factor specified on the table and metrics calculated across all interpolated frames.  We highlight the best result in \textbf{bold} and the second best with \underline{underline}.}
\label{tab:varying-scales}
\resizebox{\textwidth}{!}{%
\begin{tabular}{c|c|c|c|c|c|c|>{\columncolor{lightgrayrow}}c}
\hline
Temporal & Spatial & EMA-VFI & \multirow{2}{*}{VideoINR} & \multirow{2}{*}{MoTIF} & \multirow{2}{*}{BF-STVSR} &  \multirow{2}{*}{V$^3$} &   \\
Scale & Scale & + LIIF & & && &\multirow{-2}{*}{Ours}  \\
\hline\hline
\multirow{3}{*}{$\times 6$} & $\times 4$ &  30.64 / 0.8850 & 30.78 / 0.8954 & 31.56 / 0.9064 & 31.70 / 0.9083 & \underline{33.79} / \underline{0.9394} & \textbf{34.33} / \textbf{0.9460}\\
& $\times 6$ &  28.17 / 0.8126 & 25.56 / 0.7671 & 29.36 / 0.8505 & 29.45 / 0.8520 & \underline{30.99} / \underline{0.8867} & \textbf{31.34} / \textbf{0.8951}\\
& $\times 12$ &  24.85 / 0.7052 &  24.02 / 0.6900 & 25.81 / 0.7330 & 25.80 / 0.7295 & \underline{26.66} / \underline{0.7538} & \textbf{26.80} / \textbf{0.7589}\\
\hline
\multirow{3}{*}{$\times 12$} & $\times 4$ &  27.90 / 0.8263 &  27.32 / 0.8141 & 27.77 / 0.8230 & 28.07 / 0.8287 & \underline{28.77} / \underline{0.8432} & \textbf{29.12} / \textbf{0.8540}\\

& $\times 6$ &  26.49 / 0.7748 & 24.68 / 0.7358 & 26.78 / 0.7908 & 27.07 / 0.7963 &\underline{27.79} / \underline{0.8160} &\textbf{28.08} / \textbf{0.8276}\\
& $\times 12$ & 24.16 / 0.6918 & 23.70 / 0.6830 & 24.72 / 0.7108 & 24.88 / 0.7104 & \underline{25.56} / \underline{0.7301} & \textbf{25.74} / \textbf{0.7381}\\
\hline
\multirow{3}{*}{$\times 16$} & $\times 4$ &  26.56 / \underline{0.7904} & 25.81 / 0.7739 & 25.98 / 0.7758 & 26.39 / 0.7840  & \underline{26.68} / 0.7837 & \textbf{27.10} / \textbf{0.7993}\\
& $\times 6$ &  25.54 / 0.7503 & 23.86 / 0.7123 & 25.34 / 0.7527 & 25.81 / 0.7619 & \underline{26.14} / \underline{0.7675} & \textbf{26.40} / \textbf{0.7804}\\
& $\times 12$ &  23.68 / 0.6828 & 22.88 / 0.6659 & 23.88 / 0.6923 & 24.22 / 0.6955 & \underline{24.67} / \underline{0.7084} & \textbf{24.79} /   \textbf{0.7160}\\
\hline
\end{tabular}%
}
\end{table} \vspace{0.1cm}\\
\textbf{Arbitrary spatial and temporal scales.}
  On arbitrary-scale video SR (Appendix~\ref{app:decoupled}), our method outperforms all C-STVSR methods in both PSNR and SSIM at
  every scale from $\times2$ to $\times8$, and surpasses the dedicated ST-AVSR~\citep{shangavsr} in
  PSNR at $\times2$ and $\times8$ despite being trained on another dataset. In the temporal-only
  setting (Appendix~\ref{app:vfi}), it leads V$^3$ by 2.4\,dB PSNR and,
  on the synthesised frames alone, beats the dedicated interpolators RIFE~\citep{huang2022} and
  EMA-VFI~\citep{ema-vfi} by at least 2.2\,dB. \vspace{0.1cm}\\
\textbf{Robustness to degradations.}
Appendix~\ref{app:degraded} evaluates inputs corrupted by Gaussian noise, and by noise followed by
H.264 compression, although no method is trained on degraded inputs. Our method outperforms all C-STVSR methods at every
noise level and loses only $0.15$\,dB up to $\sigma{=}0.01$, against $0.5$\,dB for V$^3$. Under
compression, all methods are on par.\vspace{0.1cm}\\
\textbf{Qualitative results.}
Figures~\ref{fig:IMG_0015} and~\ref{fig:IMG_0004a} show frames from the Adobe240 test set at
intermediate times between two input frames, none of which the model sees. Under fast motion
(Figure~\ref{fig:IMG_0015}), prior methods either duplicate the moving object or blur it; under
parallax (Figure~\ref{fig:IMG_0004a}), they lose a thin background object seen through the
foreground. Our method keeps both sharp and at the position shown in the ground truth.
Appendix~\ref{app:qualitative} shows a fast-moving bus, where the other methods also lose thin
structures such as a street light and power lines (Figure~\ref{fig:IMG_0012}). A video comparing the different methods on the $\times4$ spatial $\times 8$ temporal task is provided in the supplementary material.
\ifdefined\stcropw\else\newlength\stcropw\newlength\stcroph\newlength\stgroupw\fi
\def\stgap{2pt}\def\stpairgap{0.5pt}
\begin{figure*}[t]
  \centering
  \begin{minipage}[c]{0.24\linewidth}
    \centering
    \includegraphics[width=\linewidth]{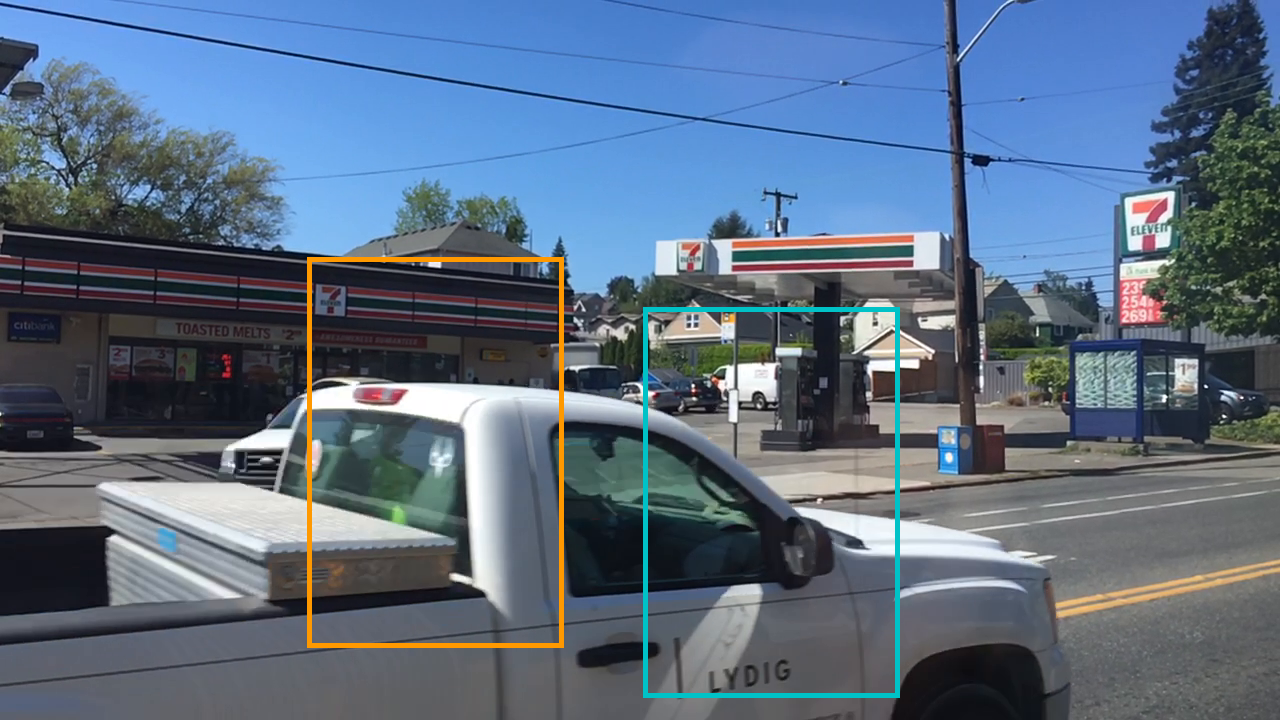}\par
    \vspace{2pt}{\small GT, frame 1565}
  \end{minipage}\hfill
  \begin{minipage}[c]{\dimexpr0.76\linewidth-6pt\relax}
    \centering
    \setlength{\fboxrule}{0.6pt}\setlength{\fboxsep}{0pt}%
    \setlength{\stcropw}{\dimexpr(\linewidth-1.1em-20\fboxrule
      -5\dimexpr\stpairgap\relax-4\dimexpr\stgap\relax-2pt)/10\relax}%
    \setlength{\stcroph}{\dimexpr\stcropw*390/256\relax}%
    \ifdim\stcroph>0.09\textheight
      \setlength{\stcroph}{0.09\textheight}\setlength{\stcropw}{\dimexpr\stcroph*256/390\relax}%
    \fi
    \setlength{\stgroupw}{\dimexpr2\stcropw+4\fboxrule+\stpairgap\relax}%
    \def\stcrop#1#2{\fcolorbox[HTML]{#1}{FFFFFF}%
      {\includegraphics[width=\stcropw,height=\stcroph]{#2}}}%
    \def\stpair#1#2{\stcrop{FF9900}{figures/IMG_0015/crops/#1/roi0/#2.png}\hspace{\stpairgap}%
      \stcrop{00C2C7}{figures/IMG_0015/crops/#1/roi1/#2.png}}%
    \def\stlabel#1{\makebox[1.1em][l]{\rotatebox{90}{\makebox[\stcroph][c]{\small #1}}}}%
    \setlength{\baselineskip}{0pt}\setlength{\lineskip}{1.5pt}\setlength{\lineskiplimit}{0pt}%
    \makebox[1.1em]{}%
    \makebox[\stgroupw][c]{\footnotesize MoTIF}\hspace{\stgap}%
    \makebox[\stgroupw][c]{\footnotesize BF-STVSR}\hspace{\stgap}%
    \makebox[\stgroupw][c]{\footnotesize $V^3$}\hspace{\stgap}%
    \makebox[\stgroupw][c]{\footnotesize Ours}\hspace{\stgap}%
    \makebox[\stgroupw][c]{\footnotesize GT}\par
    \stlabel{$\tau{=}\tfrac14$}%
    \stpair{motif}{1563}\hspace{\stgap}\stpair{bfstvsr}{1563}\hspace{\stgap}%
    \stpair{v3}{1563}\hspace{\stgap}\stpair{Ours}{1563}\hspace{\stgap}\stpair{GT}{1563}\par
    \stlabel{$\tau{=}\tfrac12$}%
    \stpair{motif}{1565}\hspace{\stgap}\stpair{bfstvsr}{1565}\hspace{\stgap}%
    \stpair{v3}{1565}\hspace{\stgap}\stpair{Ours}{1565}\hspace{\stgap}\stpair{GT}{1565}\par
    \stlabel{$\tau{=}\tfrac34$}%
    \stpair{motif}{1567}\hspace{\stgap}\stpair{bfstvsr}{1567}\hspace{\stgap}%
    \stpair{v3}{1567}\hspace{\stgap}\stpair{Ours}{1567}\hspace{\stgap}\stpair{GT}{1567}\par
  \end{minipage}
  \caption{Fast motion on Adobe240 IMG\_0015 ($\times$4 space, $\times$8 time), at the intermediate
times between input keyframes 1561 and 1569. A truck drives across the frame. Ours keeps the back of
the truck sharp and the ``LYDIG'' text on its door readable at every intermediate time. $V^3$ blurs
the truck and adds straight lines where the borders between its space-time processing blocks show
in the output, while MoTIF and BF-STVSR
produce several overlapping copies of it. Recovered details (or lack thereof) may be better appreciated by zooming on the figure.}
  \label{fig:IMG_0015}
\end{figure*}
\ifdefined\stcropw\else\newlength\stcropw\newlength\stcroph\newlength\stgroupw\fi
\begin{figure*}[t]
  \centering
  \begin{minipage}[c]{0.24\linewidth}
    \centering
    \includegraphics[width=\linewidth]{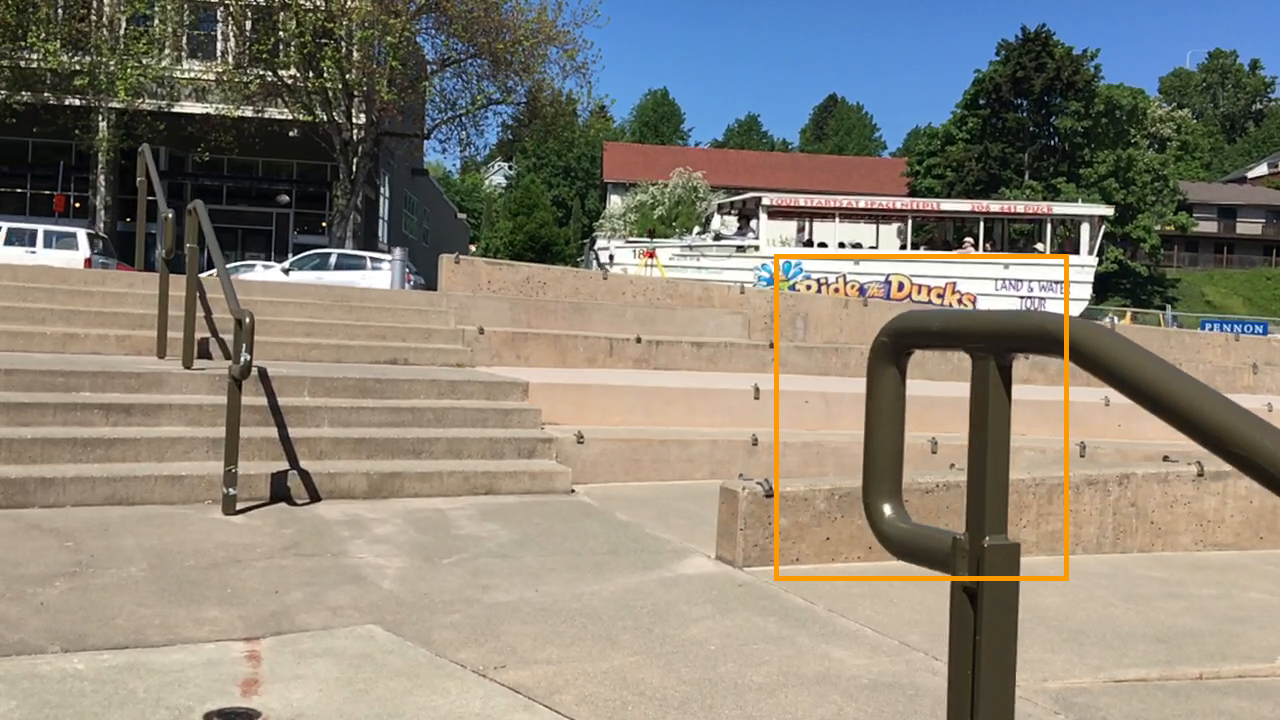}\par
    \vspace{2pt}{\small GT, frame 237}
  \end{minipage}\hfill
  \begin{minipage}[c]{\dimexpr0.76\linewidth-6pt\relax}
    \centering
    \setlength{\fboxrule}{0.6pt}\setlength{\fboxsep}{0pt}%
    \setlength{\stcropw}{\dimexpr(\linewidth-1.1em-10\fboxrule-4\dimexpr4pt\relax-2pt)/5\relax}%
    \setlength{\stcroph}{\dimexpr\stcropw*326/294\relax}%
    \ifdim\stcroph>0.09\textheight\setlength{\stcroph}{0.09\textheight}\setlength{\stcropw}{\dimexpr\stcroph*294/326\relax}\fi
    \setlength{\stgroupw}{\dimexpr\stcropw+2\fboxrule\relax}%
    \def\stcrop#1#2{\fcolorbox[HTML]{#1}{FFFFFF}{\includegraphics[width=\stcropw,height=\stcroph]{#2}}}%
    \def\stlabel#1{\makebox[1.1em][l]{\rotatebox{90}{\makebox[\stcroph][c]{\small #1}}}}%
    \setlength{\baselineskip}{0pt}\setlength{\lineskip}{1.5pt}\setlength{\lineskiplimit}{0pt}%
    \makebox[1.1em]{}%
    \makebox[\stgroupw][c]{\footnotesize MoTIF}\hspace{4pt}%
    \makebox[\stgroupw][c]{\footnotesize BF-STVSR}\hspace{4pt}%
    \makebox[\stgroupw][c]{\footnotesize $V^3$}\hspace{4pt}%
    \makebox[\stgroupw][c]{\footnotesize Ours}\hspace{4pt}%
    \makebox[\stgroupw][c]{\footnotesize GT}\par
    \stlabel{$\tau{=}\tfrac14$}%
    \stcrop{FF9900}{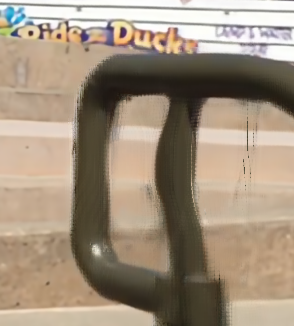}\hspace{4pt}%
    \stcrop{FF9900}{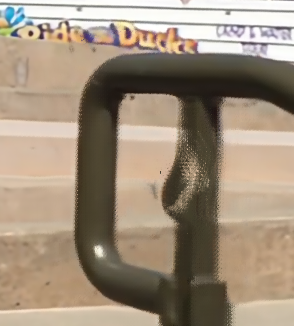}\hspace{4pt}%
    \stcrop{FF9900}{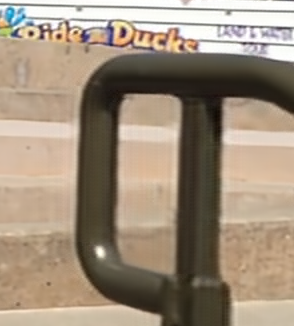}\hspace{4pt}%
    \stcrop{FF9900}{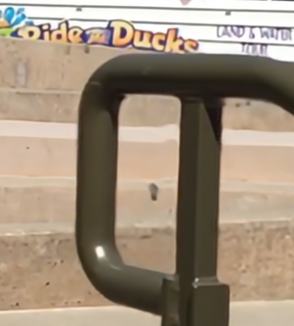}\hspace{4pt}%
    \stcrop{FF9900}{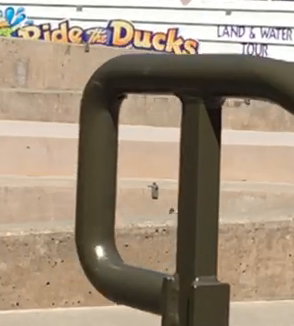}\par
    \stlabel{$\tau{=}\tfrac12$}%
    \stcrop{FF9900}{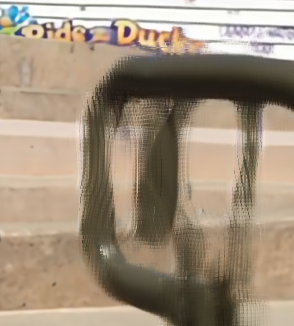}\hspace{4pt}%
    \stcrop{FF9900}{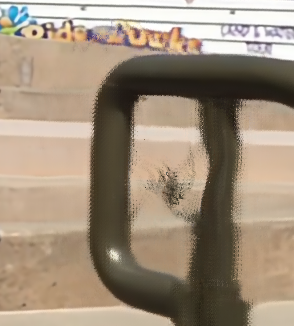}\hspace{4pt}%
    \stcrop{FF9900}{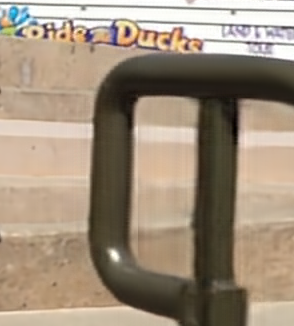}\hspace{4pt}%
    \stcrop{FF9900}{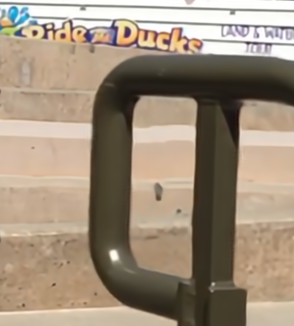}\hspace{4pt}%
    \stcrop{FF9900}{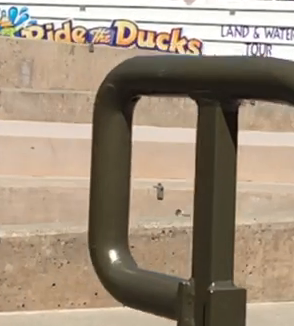}\par
    \stlabel{$\tau{=}\tfrac34$}%
    \stcrop{FF9900}{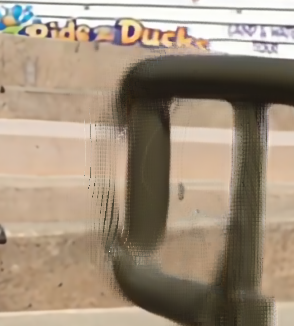}\hspace{4pt}%
    \stcrop{FF9900}{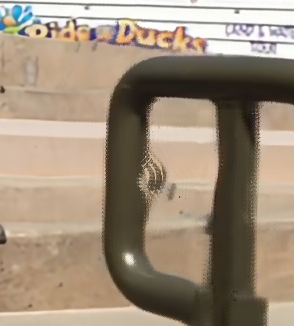}\hspace{4pt}%
    \stcrop{FF9900}{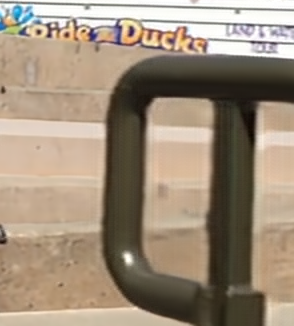}\hspace{4pt}%
    \stcrop{FF9900}{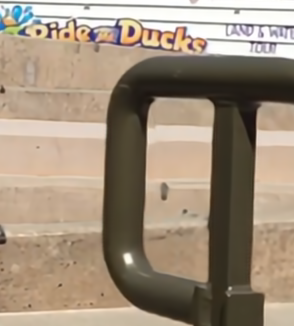}\hspace{4pt}%
    \stcrop{FF9900}{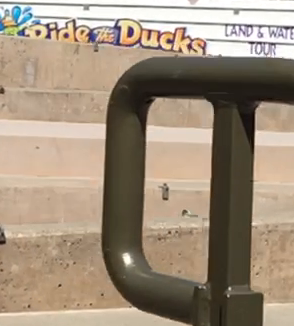}\par
  \end{minipage}
\caption{Parallax on Adobe240 IMG\_0004a ($\times$4 space, $\times$8 time), at the intermediate times
between input keyframes 233 and 241. The scene pans around with a rail in the foreground and a staircase in the background is visible through the railing loop. Our method keeps the bar intact and correctly pans the background at every intermediate time. $V^3$ blurs the rail and background through the loop, BF-STVSR breaks the bar where it overlaps the background, and
MoTIF smears the two layers together. Recovered details (or lack thereof) may be better appreciated by zooming on the figure.}
  \label{fig:IMG_0004a}
\end{figure*}

\section{Conclusion}
\textbf{Discussion.} We showed that the principle of classical multi-frame super-resolution,
accumulating several frames at their sub-pixel positions on a finer grid, carries over to learned
continuous space-time video super-resolution. What we splat, however, are not raw observations but
features that a video encoder has propagated across the whole sequence, so each splatted sample
already carries temporal context beyond its own frame before a decoder maps the fused features to
pixels. This combines the two paradigms that learned video super-resolution had separated: implicit
feature propagation supplies temporal context, while explicit sub-pixel accumulation keeps each
sample at its own position until frames are fused. As a result, a single splat both interpolates in
time and gathers detail from neighbouring frames, improving fidelity and perceptual quality at a
fraction of the computational cost. Natural next steps are a visibility head decoupled from the
motion basis, and more samples per target pixel, by splatting more frames or several trajectories
per pixel~\mbox{\citep{manytomany}}, to go beyond our fixed $\times4$ splat, which is limited by the two
frames it combines \mbox{\citep{baker2002, wronski2019handheld}}. Edge-aligned splat
kernels~\citep{wronski2019handheld} could also fill holes without blurring sub-pixel detail.\vspace{0.1cm}\\
\textbf{Limitations.} (1)~Visibility is read from the trajectory's B-spline basis, following
BF-STVSR~\citep{bfstvsr}, so it is as smooth in time as the motion and tied to the point's own
trajectory rather than to its occluder. This limits its gain (Appendix~\ref{app:visibility}) and
leaves errors at the boundaries of moving objects. (2)~Under complex motion, trajectories can
place content a few pixels off (Appendix~\ref{app:qualitative}), which costs PSNR even when the
frame is sharp. (3)~Like all evaluated
methods, ours does not recover detail from compressed inputs (Appendix~\ref{app:degraded}).\vspace{0.1cm}\\

\section{Acknowledgements}
J.M. was supported by the European Research Council (ERC) under grant agreements 101087696 (APHELEIA project). 
J.P. was supported in part by the French government under management of Agence Nationale de la Recherche as part of the “France 2030” program, PR[AI]RIE-PSAI project, reference ANR-23- IACL-0008, the Institute of Information \& Communications Technology Planning \& Evaluation (IITP) grant funded by the Korean Government (MSIT) (No. RS-2024-00457882, National AI Research Lab Project), and a Global Distinguished Professorship at the Courant Institute of Mathematical Sciences and the Center for Data Science at New York University.
This work was granted
access to the HPC resources of IDRIS under the allocation AD011016624R1 made by GENCI.

\section*{AI use statement}

The research direction, method, experiments and conclusions of this work are the authors' own.
We used generative AI tools as a discussion partner to help organise and refine our ideas, to suggest experiments and ablations, which we then designed, ran and checked ourselves; every reported result comes from these experiments. We
also used them to help implement parts of the code, to search for related work, to obtain feedback
on the manuscript, and to draft and edit parts of the text. All AI-assisted code was reviewed and
tested by the authors, and all text was checked and revised by the authors. The remaining tasks
listed as requiring disclosure (synthetic data generation, mathematical claims or proofs, dataset
cleaning, translation) are not applicable. We take responsibility for the final content of this
work.

\section*{Ethics Statement}
This work uses only publicly available video benchmarks (Adobe240, GoPro, REDS and Vid4),
under their respective licenses and terms of use. No new data was collected and no human
subjects were involved. While spatial detail is recovered by aggregating sub-pixel
observations across neighbouring frames, intermediate frames are synthesised at time
stamps that were never captured. The motion they depict follows a learned trajectory
model and is not observed. In applications where frame content carries evidential
weight, such as forensic, medical or surveillance analysis, synthesised intermediate
frames should be clearly distinguished from captured ones and interpreted with
appropriate validation.

\section*{Reproducibility Statement}
  The method is described in Section~\ref{sec:method}: the encoder in Section~\ref{sec:encoder},
  the trajectory and visibility estimation in Section~\ref{sec:trajectory}, with the B-spline
  construction in Appendix~\ref{app:b-splines}, the splatting in Section~\ref{sec:splatting} and the
  decoder in Section~\ref{sec:decoding}. Architecture hyperparameters, the training objective and
  schedule, the hardware and the dataset splits are given in Section~\ref{sec:setup}. The evaluation
  protocol for each benchmark is given with its results: the \emph{Center} and \emph{Average} subsets
  in Section~\ref{sec:cstvsr-results}, arbitrary-scale SR in Appendix~\ref{app:decoupled}, frame
  interpolation in Appendix~\ref{app:vfi}, the noise and compression settings, including the codec
  parameters, in Appendix~\ref{app:degraded}, the additional metrics and the VMAF encoding in
  Appendix~\ref{app:additional-metrics}, and runtime measurement in Appendix~\ref{app:runtime}. The inference-time trajectory study is described in Appendix~\ref{app:traj-frames}, and the
visibility ablation, including the definition of the occluded areas, in
Appendix~\ref{app:visibility}. We re-evaluated every baseline with public weights using its official code under this protocol;
  where our results matched the published ones, we report the published values.

\bibliographystyle{plain}
\bibliography{references}

\begin{thebibliography}{60}
\providecommand{\natexlab}[1]{#1}
\providecommand{\url}[1]{\texttt{#1}}
\expandafter\ifx\csname urlstyle\endcsname\relax
  \providecommand{\doi}[1]{doi: #1}\else
  \providecommand{\doi}{doi: \begingroup \urlstyle{rm}\Url}\fi

\bibitem[Baker \& Kanade(2002)Baker and Kanade]{baker2002}
Simon Baker and Takeo Kanade.
\newblock Limits on super-resolution and how to break them.
\newblock \emph{IEEE Transactions on Pattern Analysis and Machine
  Intelligence}, 24\penalty0 (9):\penalty0 1167--1183, 2002.

\bibitem[Bao et~al.(2019)Bao, Lai, Ma, Zhang, Gao, and Yang]{bao2019depth}
Wenbo Bao, Wei-Sheng Lai, Chao Ma, Xiaoyun Zhang, Zhiyong Gao, and Ming-Hsuan
  Yang.
\newblock Depth-aware video frame interpolation.
\newblock In \emph{IEEE/CVF Conference on Computer Vision and Pattern
  Recognition (CVPR)}, 2019.

\bibitem[Becker et~al.(2026)Becker, Erbach, Narnhofer, and Schindler]{v-cube}
Alexander Becker, Julius Erbach, Dominik Narnhofer, and Konrad Schindler.
\newblock Continuous space-time video super-resolution with {3D} {Fourier}
  fields.
\newblock In \emph{International Conference on Learning Representations
  (ICLR)}, 2026.

\bibitem[Chan et~al.(2021{\natexlab{a}})Chan, Wang, Yu, Dong, and
  Loy]{basicvsr}
Kelvin C.~K. Chan, Xintao Wang, Ke~Yu, Chao Dong, and Chen~Change Loy.
\newblock {BasicVSR}: The search for essential components in video
  super-resolution and beyond.
\newblock In \emph{IEEE/CVF Conference on Computer Vision and Pattern
  Recognition (CVPR)}, 2021{\natexlab{a}}.

\bibitem[Chan et~al.(2021{\natexlab{b}})Chan, Wang, Yu, Dong, and
  Loy]{deepdiveDCN}
Kelvin C.~K. Chan, Xintao Wang, Ke~Yu, Chao Dong, and Chen~Change Loy.
\newblock Understanding deformable alignment in video super-resolution.
\newblock In \emph{AAAI Conference on Artificial Intelligence (AAAI)},
  2021{\natexlab{b}}.

\bibitem[Chan et~al.(2022{\natexlab{a}})Chan, Zhou, Xu, and Loy]{basicvsrpp}
Kelvin C.~K. Chan, Shangchen Zhou, Xiangyu Xu, and Chen~Change Loy.
\newblock {BasicVSR++}: Improving video super-resolution with enhanced
  propagation and alignment.
\newblock In \emph{IEEE/CVF Conference on Computer Vision and Pattern
  Recognition (CVPR)}, 2022{\natexlab{a}}.

\bibitem[Chan et~al.(2022{\natexlab{b}})Chan, Zhou, Xu, and Loy]{realbasicvsr}
Kelvin C.~K. Chan, Shangchen Zhou, Xiangyu Xu, and Chen~Change Loy.
\newblock Investigating tradeoffs in real-world video super-resolution,
  2022{\natexlab{b}}.

\bibitem[Chen et~al.(2025)Chen, Chen, Zhang, and Zhang]{gsasr}
Du~Chen, Liyi Chen, Zhengqiang Zhang, and Lei Zhang.
\newblock Generalized and efficient {2D} {Gaussian} splatting for
  arbitrary-scale super-resolution.
\newblock In \emph{IEEE/CVF International Conference on Computer Vision
  (ICCV)}, 2025.

\bibitem[Chen et~al.(2023{\natexlab{a}})Chen, Xu, Hong, Tsai, Kuo, and
  Lee]{clit}
Hao-Wei Chen, Yu-Syuan Xu, Min-Fong Hong, Yi-Min Tsai, Hsien-Kai Kuo, and
  Chun-Yi Lee.
\newblock Cascaded local implicit transformer for arbitrary-scale
  super-resolution.
\newblock In \emph{IEEE/CVF Conference on Computer Vision and Pattern
  Recognition (CVPR)}, 2023{\natexlab{a}}.

\bibitem[Chen et~al.(2023{\natexlab{b}})Chen, Chen, Chen, Lin, and Peng]{motif}
Yi-Hsin Chen, Si-Cun Chen, Yi-Hsin Chen, Yen-Yu Lin, and Wen-Hsiao Peng.
\newblock {MoTIF}: Learning motion trajectories with local implicit neural
  functions for continuous space-time video super-resolution.
\newblock In \emph{IEEE/CVF International Conference on Computer Vision
  (ICCV)}, 2023{\natexlab{b}}.

\bibitem[Chen et~al.(2021)Chen, Liu, and Wang]{liif}
Yinbo Chen, Sifei Liu, and Xiaolong Wang.
\newblock Learning continuous image representation with local implicit image
  function.
\newblock In \emph{IEEE/CVF Conference on Computer Vision and Pattern
  Recognition (CVPR)}, 2021.

\bibitem[Chen et~al.(2022)Chen, Chen, Liu, Xu, Goel, Wang, Shi, and
  Wang]{videoinr}
Zeyuan Chen, Yinbo Chen, Jingwen Liu, Xingqian Xu, Vidit Goel, Zhangyang Wang,
  Humphrey Shi, and Xiaolong Wang.
\newblock {VideoINR}: Learning video implicit neural representation for
  continuous space-time super-resolution.
\newblock In \emph{IEEE/CVF Conference on Computer Vision and Pattern
  Recognition (CVPR)}, 2022.

\bibitem[Chu et~al.(2020)Chu, Xie, Mayer, Leal-Taix{\'e}, and Thuerey]{tof}
Mengyu Chu, You Xie, Jonas Mayer, Laura Leal-Taix{\'e}, and Nils Thuerey.
\newblock Learning temporal coherence via self-supervision for {GAN}-based
  video generation.
\newblock \emph{ACM Transactions on Graphics}, 39\penalty0 (4):\penalty0
  75:1--75:13, 2020.

\bibitem[Danier et~al.(2022)Danier, Zhang, and Bull]{flolpips}
Duolikun Danier, Fan Zhang, and David Bull.
\newblock {FloLPIPS}: A bespoke video quality metric for frame interpolation.
\newblock In \emph{Picture Coding Symposium (PCS)}, 2022.

\bibitem[Farsiu et~al.(2004)Farsiu, Robinson, Elad, and Milanfar]{farsiu2004}
Sina Farsiu, M.~Dirk Robinson, Michael Elad, and Peyman Milanfar.
\newblock Fast and robust multiframe super resolution.
\newblock \emph{IEEE Transactions on Image Processing}, 13\penalty0
  (10):\penalty0 1327--1344, 2004.

\bibitem[Haris et~al.(2020)Haris, Shakhnarovich, and Ukita]{haris2020spacetime}
Muhammad Haris, Greg Shakhnarovich, and Norimichi Ukita.
\newblock Space-time-aware multi-resolution video enhancement.
\newblock In \emph{IEEE/CVF Conference on Computer Vision and Pattern
  Recognition (CVPR)}, 2020.

\bibitem[Hou et~al.(2022)Hou, Ghildyal, and Liu]{vfips}
Qiqi Hou, Abhijay Ghildyal, and Feng Liu.
\newblock A perceptual quality metric for video frame interpolation.
\newblock In \emph{European Conference on Computer Vision (ECCV)}, 2022.

\bibitem[Hu et~al.(2022)Hu, Niklaus, Sclaroff, and Saenko]{manytomany}
Ping Hu, Simon Niklaus, Stan Sclaroff, and Kate Saenko.
\newblock Many-to-many splatting for efficient video frame interpolation.
\newblock In \emph{IEEE/CVF Conference on Computer Vision and Pattern
  Recognition (CVPR)}, 2022.

\bibitem[Huang et~al.(2022)Huang, Zhang, Heng, Shi, and Zhou]{huang2022}
Zhewei Huang, Tianyuan Zhang, Wen Heng, Boxin Shi, and Shuchang Zhou.
\newblock Real-time intermediate flow estimation for video frame interpolation.
\newblock In \emph{European Conference on Computer Vision (ECCV)}, 2022.

\bibitem[Irani \& Peleg(1991)Irani and Peleg]{iranipeleg1991}
Michal Irani and Shmuel Peleg.
\newblock Improving resolution by image registration.
\newblock \emph{CVGIP: Graphical Models and Image Processing}, 53\penalty0
  (3):\penalty0 231--239, 1991.

\bibitem[Jiang et~al.(2018)Jiang, Sun, Jampani, Yang, Learned-Miller, and
  Kautz]{superslomo}
Huaizu Jiang, Deqing Sun, Varun Jampani, Ming-Hsuan Yang, Erik Learned-Miller,
  and Jan Kautz.
\newblock Super {SloMo}: High quality estimation of multiple intermediate
  frames for video interpolation.
\newblock In \emph{IEEE/CVF Conference on Computer Vision and Pattern
  Recognition (CVPR)}, 2018.

\bibitem[Jiang et~al.(2025)Jiang, Kwan, Peng, Zhu, Sole, and Bull]{hiif}
Yuxuan Jiang, Ho~Man Kwan, Tianhao Peng, Xiaoqing Zhu, Joel Sole, and David
  Bull.
\newblock {HIIF}: Hierarchical encoding based implicit image function for
  continuous super-resolution.
\newblock In \emph{IEEE/CVF Conference on Computer Vision and Pattern
  Recognition (CVPR)}, 2025.

\bibitem[Kim et~al.(2025)Kim, Kim, Jin, and Yoo]{bfstvsr}
Eunjin Kim, Hyeonjin Kim, Kyong~Hwan Jin, and Jaejun Yoo.
\newblock {BF-STVSR}: {B-Splines} and {Fourier}---best friends for high
  fidelity spatial-temporal video super-resolution.
\newblock In \emph{IEEE/CVF Conference on Computer Vision and Pattern
  Recognition (CVPR)}, 2025.

\bibitem[Lai et~al.(2017)Lai, Huang, Ahuja, and Yang]{charb}
Wei-Sheng Lai, Jia-Bin Huang, Narendra Ahuja, and Ming-Hsuan Yang.
\newblock Deep {Laplacian} pyramid networks for fast and accurate
  super-resolution.
\newblock In \emph{IEEE/CVF Conference on Computer Vision and Pattern
  Recognition (CVPR)}, 2017.

\bibitem[Lee \& Jin(2022)Lee and Jin]{lte}
Jaewon Lee and Kyong~Hwan Jin.
\newblock Local texture estimator for implicit representation function.
\newblock In \emph{IEEE/CVF Conference on Computer Vision and Pattern
  Recognition (CVPR)}, pp.\  1929--1938, 2022.

\bibitem[Li et~al.(2021)Li, Jin, Yang, Liu, Yang, and Milanfar]{comisr}
Yinxiao Li, Pengchong Jin, Feng Yang, Ce~Liu, Ming-Hsuan Yang, and Peyman
  Milanfar.
\newblock Comisr: Compression-informed video super-resolution.
\newblock In \emph{IEEE/CVF International Conference on Computer Vision
  (ICCV)}, 2021.

\bibitem[Li et~al.(2024)Li, Liu, Shang, Liu, Wan, and Feng]{liavsr}
Zekun Li, Hongying Liu, Fanhua Shang, Yuanyuan Liu, Liang Wan, and Wei Feng.
\newblock {SAVSR}: Arbitrary-scale video super-resolution via a learned
  scale-adaptive network.
\newblock In \emph{AAAI Conference on Artificial Intelligence (AAAI)}, 2024.

\bibitem[Li et~al.(2016)Li, Aaron, Katsavounidis, Moorthy, and Manohara]{vmaf}
Zhi Li, Anne Aaron, Ioannis Katsavounidis, Anush Moorthy, and Megha Manohara.
\newblock Toward a practical perceptual video quality metric.
\newblock Netflix Technology Blog, 2016.
\newblock URL
  \url{https://netflixtechblog.com/toward-a-practical-perceptual-video-quality-metric-653f208b9652}.

\bibitem[Liang et~al.(2022)Liang, Fan, Xiang, Ranjan, Ilg, Green, Cao, Zhang,
  Timofte, and Van~Gool]{rvrt}
Jingyun Liang, Yuchen Fan, Xiaoyu Xiang, Rakesh Ranjan, Eddy Ilg, Simon Green,
  Jiezhang Cao, Kai Zhang, Radu Timofte, and Luc Van~Gool.
\newblock Recurrent video restoration transformer with guided deformable
  attention.
\newblock In \emph{Advances in Neural Information Processing Systems
  (NeurIPS)}, 2022.

\bibitem[Lin \& Shum(2004)Lin and Shum]{lin2004}
Zhouchen Lin and Heung-Yeung Shum.
\newblock Fundamental limits of reconstruction-based superresolution algorithms
  under local translation.
\newblock \emph{IEEE Transactions on Pattern Analysis and Machine
  Intelligence}, 26\penalty0 (1):\penalty0 83--97, 2004.

\bibitem[Liu \& Sun(2011)Liu and Sun]{viddataset}
Ce~Liu and Deqing Sun.
\newblock A {Bayesian} approach to adaptive video super resolution.
\newblock In \emph{IEEE/CVF Conference on Computer Vision and Pattern
  Recognition (CVPR)}, 2011.

\bibitem[Liu et~al.(2020)Liu, Xie, Siyao, Sun, Qiao, and Dong]{eqvi}
Yihao Liu, Liangbin Xie, Li~Siyao, Wenxiu Sun, Yu~Qiao, and Chao Dong.
\newblock Enhanced quadratic video interpolation.
\newblock In \emph{European Conference on Computer Vision Workshops (ECCVW)},
  2020.

\bibitem[Nah et~al.(2017)Nah, Kim, and Lee]{gopro}
Seungjun Nah, Tae~Hyun Kim, and Kyoung~Mu Lee.
\newblock Deep multi-scale convolutional neural network for dynamic scene
  deblurring.
\newblock In \emph{IEEE/CVF Conference on Computer Vision and Pattern
  Recognition (CVPR)}, 2017.

\bibitem[Nah et~al.(2019)Nah, Baik, Hong, Moon, Son, Timofte, and Lee]{reds}
Seungjun Nah, Sungyong Baik, Seokil Hong, Gyeongsik Moon, Sanghyun Son, Radu
  Timofte, and Kyoung~Mu Lee.
\newblock {NTIRE} 2019 challenge on video deblurring and super-resolution:
  Dataset and study.
\newblock In \emph{IEEE/CVF Conference on Computer Vision and Pattern
  Recognition Workshops (CVPRW)}, 2019.

\bibitem[Niklaus \& Liu(2020)Niklaus and Liu]{softsplat}
Simon Niklaus and Feng Liu.
\newblock Softmax splatting for video frame interpolation.
\newblock In \emph{IEEE/CVF Conference on Computer Vision and Pattern
  Recognition (CVPR)}, 2020.

\bibitem[Niklaus et~al.(2017)Niklaus, Mai, and Liu]{niklaus2017video}
Simon Niklaus, Long Mai, and Feng Liu.
\newblock Video frame interpolation via adaptive separable convolution.
\newblock In \emph{IEEE/CVF International Conference on Computer Vision
  (ICCV)}, 2017.

\bibitem[Pak et~al.(2023)Pak, Lee, and Jin]{pak2023b}
Byeonghyun Pak, Jaewon Lee, and Kyong~Hwan Jin.
\newblock {B-Spline} texture coefficients estimator for screen content image
  super-resolution.
\newblock In \emph{IEEE/CVF Conference on Computer Vision and Pattern
  Recognition (CVPR)}, 2023.

\bibitem[Peng et~al.(2025)Peng, Wu, Li, Xia, Dai, Zhang, Sun, Pei, Wang, Cao,
  and Zha]{pixel2gaussian}
Long Peng, Anran Wu, Wenbo Li, Peizhe Xia, Xueyuan Dai, Xinjie Zhang, Haoze
  Sun, Renjing Pei, Yang Wang, Yang Cao, and Zheng-Jun Zha.
\newblock Pixel to {Gaussian}: Ultra-fast continuous super-resolution with {2D}
  {Gaussian} modeling.
\newblock In \emph{International Conference on Learning Representations
  (ICLR)}, 2025.

\bibitem[Shang et~al.(2024)Shang, Ren, Zhang, Fang, Zuo, and Ma]{shangavsr}
Wei Shang, Dongwei Ren, Wanying Zhang, Yuming Fang, Wangmeng Zuo, and Kede Ma.
\newblock Arbitrary-scale video super-resolution with structural and textural
  priors.
\newblock In \emph{European Conference on Computer Vision (ECCV)}, 2024.

\bibitem[Shi et~al.(2026)Shi, Di, Peng, Cao, Wu, Feng, Guo, Pei, Fu, Cao, and
  Zha]{gsstvsr}
Mingyu Shi, Xin Di, Long Peng, Boxiang Cao, Anran Wu, Zhanfeng Feng, Jiaming
  Guo, Renjing Pei, Xueyang Fu, Yang Cao, and Zheng-Jun Zha.
\newblock {GS-STVSR}: Ultra-efficient continuous spatio-temporal video
  super-resolution via {2D} {Gaussian} splatting.
\newblock \emph{arXiv preprint arXiv:2604.18047}, 2026.

\bibitem[Shi et~al.(2022)Shi, Gu, Xie, Wang, Yang, and Dong]{psrt}
Shuwei Shi, Jinjin Gu, Liangbin Xie, Xintao Wang, Yujiu Yang, and Chao Dong.
\newblock Rethinking alignment in video super-resolution transformers.
\newblock In \emph{Advances in Neural Information Processing Systems
  (NeurIPS)}, 2022.

\bibitem[Shi et~al.(2016)Shi, Caballero, Husz{\'a}r, Totz, Aitken, Bishop,
  Rueckert, and Wang]{shi2016real}
Wenzhe Shi, Jose Caballero, Ferenc Husz{\'a}r, Johannes Totz, Andrew~P. Aitken,
  Rob Bishop, Daniel Rueckert, and Zehan Wang.
\newblock Real-time single image and video super-resolution using an efficient
  sub-pixel convolutional neural network.
\newblock In \emph{IEEE/CVF Conference on Computer Vision and Pattern
  Recognition (CVPR)}, 2016.

\bibitem[Sim et~al.(2021)Sim, Oh, and Kim]{sim2021xvfiextremevideoframe}
Hyeonjun Sim, Jihyong Oh, and Munchurl Kim.
\newblock {XVFI}: {eXtreme} video frame interpolation.
\newblock In \emph{IEEE/CVF International Conference on Computer Vision
  (ICCV)}, 2021.

\bibitem[Sitzmann et~al.(2020)Sitzmann, Martel, Bergman, Lindell, and
  Wetzstein]{inrper}
Vincent Sitzmann, Julien N.~P. Martel, Alexander~W. Bergman, David~B. Lindell,
  and Gordon Wetzstein.
\newblock Implicit neural representations with periodic activation functions.
\newblock In \emph{Advances in Neural Information Processing Systems
  (NeurIPS)}, 2020.

\bibitem[Su et~al.(2017)Su, Delbracio, Wang, Sapiro, Heidrich, and
  Wang]{adobe240}
Shuochen Su, Mauricio Delbracio, Jue Wang, Guillermo Sapiro, Wolfgang Heidrich,
  and Oliver Wang.
\newblock Deep video deblurring for hand-held cameras.
\newblock In \emph{IEEE/CVF Conference on Computer Vision and Pattern
  Recognition (CVPR)}, 2017.

\bibitem[Tao et~al.(2017)Tao, Gao, Liao, Wang, and Jia]{spmc}
Xin Tao, Hongyun Gao, Renjie Liao, Jue Wang, and Jiaya Jia.
\newblock Detail-revealing deep video super-resolution.
\newblock In \emph{IEEE/CVF International Conference on Computer Vision
  (ICCV)}, 2017.

\bibitem[Teed \& Deng(2020)Teed and Deng]{raft}
Zachary Teed and Jia Deng.
\newblock {RAFT}: Recurrent all-pairs field transforms for optical flow.
\newblock In \emph{European Conference on Computer Vision (ECCV)}, 2020.

\bibitem[Tran et~al.(2025)Tran, Hung, and Kim]{vsrm}
Dinh~Phu Tran, Dao~Duy Hung, and Daeyoung Kim.
\newblock {VSRM}: A robust {Mamba}-based framework for video super-resolution.
\newblock In \emph{IEEE/CVF International Conference on Computer Vision
  (ICCV)}, 2025.

\bibitem[Tsai \& Huang(1984)Tsai and Huang]{tsaihuang1984}
Roger~Y. Tsai and Thomas~S. Huang.
\newblock Multiframe image restoration and registration.
\newblock In \emph{Advances in Computer Vision and Image Processing}, volume~1,
  pp.\  317--339. JAI Press, 1984.

\bibitem[Wang et~al.(2019)Wang, Chan, Yu, Dong, and Loy]{edvr}
Xintao Wang, Kelvin C.~K. Chan, Ke~Yu, Chao Dong, and Chen~Change Loy.
\newblock {EDVR}: Video restoration with enhanced deformable convolutional
  networks.
\newblock In \emph{IEEE/CVF Conference on Computer Vision and Pattern
  Recognition Workshops (CVPRW)}, 2019.

\bibitem[Wronski et~al.(2019)Wronski, Garcia-Dorado, Ernst, Kelly, Krainin,
  Liang, Levoy, and Milanfar]{wronski2019handheld}
Bartlomiej Wronski, Ignacio Garcia-Dorado, Manfred Ernst, Damien Kelly, Michael
  Krainin, Chia-Kai Liang, Marc Levoy, and Peyman Milanfar.
\newblock Handheld multi-frame super-resolution.
\newblock \emph{ACM Transactions on Graphics}, 38\penalty0 (4):\penalty0
  28:1--28:18, 2019.

\bibitem[Xiang et~al.(2020)Xiang, Tian, Zhang, Fu, Allebach, and
  Xu]{zoomingslowmo}
Xiaoyu Xiang, Yapeng Tian, Yulun Zhang, Yun Fu, Jan~P. Allebach, and Chenliang
  Xu.
\newblock Zooming {Slow-Mo}: Fast and accurate one-stage space-time video
  super-resolution.
\newblock In \emph{IEEE/CVF Conference on Computer Vision and Pattern
  Recognition (CVPR)}, 2020.

\bibitem[Xu et~al.(2021)Xu, Xu, Li, Wang, Sun, and Cheng]{xu2021temporal}
Gang Xu, Jun Xu, Zhen Li, Liang Wang, Xing Sun, and Ming-Ming Cheng.
\newblock Temporal modulation network for controllable space-time video
  super-resolution.
\newblock In \emph{IEEE/CVF Conference on Computer Vision and Pattern
  Recognition (CVPR)}, 2021.

\bibitem[Xu et~al.(2024)Xu, Yu, Wang, Mi, and Yao]{iart}
Kai Xu, Ziwei Yu, Xin Wang, Michael~Bi Mi, and Angela Yao.
\newblock Enhancing video super-resolution via implicit resampling-based
  alignment.
\newblock In \emph{IEEE/CVF Conference on Computer Vision and Pattern
  Recognition (CVPR)}, 2024.

\bibitem[Xu et~al.(2019)Xu, Siyao, Sun, Yin, and Yang]{qvi}
Xiangyu Xu, Li~Siyao, Wenxiu Sun, Qian Yin, and Ming-Hsuan Yang.
\newblock Quadratic video interpolation.
\newblock In \emph{Advances in Neural Information Processing Systems
  (NeurIPS)}, 2019.

\bibitem[Xue et~al.(2019)Xue, Chen, Wu, Wei, and Freeman]{vimeo90k}
Tianfan Xue, Baian Chen, Jiajun Wu, Donglai Wei, and William~T. Freeman.
\newblock Video enhancement with task-oriented flow.
\newblock \emph{International Journal of Computer Vision}, 127\penalty0
  (8):\penalty0 1106--1125, 2019.

\bibitem[Zhang et~al.(2023)Zhang, Zhu, Wang, Chen, Wu, and Wang]{ema-vfi}
Guozhen Zhang, Yuhan Zhu, Haonan Wang, Youxin Chen, Gangshan Wu, and Limin
  Wang.
\newblock Extracting motion and appearance via inter-frame attention for
  efficient video frame interpolation.
\newblock In \emph{IEEE/CVF Conference on Computer Vision and Pattern
  Recognition (CVPR)}, 2023.

\bibitem[Zhang et~al.(2018{\natexlab{a}})Zhang, Isola, Efros, Shechtman, and
  Wang]{lpips}
Richard Zhang, Phillip Isola, Alexei~A. Efros, Eli Shechtman, and Oliver Wang.
\newblock The unreasonable effectiveness of deep features as a perceptual
  metric.
\newblock In \emph{IEEE/CVF Conference on Computer Vision and Pattern
  Recognition (CVPR)}, 2018{\natexlab{a}}.

\bibitem[Zhang et~al.(2018{\natexlab{b}})Zhang, Tian, Kong, Zhong, and Fu]{rdn}
Yulun Zhang, Yapeng Tian, Yu~Kong, Bineng Zhong, and Yun Fu.
\newblock Residual dense network for image super-resolution.
\newblock In \emph{IEEE/CVF Conference on Computer Vision and Pattern
  Recognition (CVPR)}, 2018{\natexlab{b}}.

\bibitem[Zhang et~al.(2018{\natexlab{c}})Zhang, Liu, and Wang]{resunet}
Zhengxin Zhang, Qingjie Liu, and Yunhong Wang.
\newblock Road extraction by deep residual {U-Net}.
\newblock \emph{IEEE Geoscience and Remote Sensing Letters}, 15\penalty0
  (5):\penalty0 749--753, 2018{\natexlab{c}}.

\end{thebibliography}

\newpage
\appendix

\section{Importance of temporal information for motion estimation}
\label{app:traj-frames}

We examine whether the trajectory estimator uses temporal context beyond the two frames
bracketing the interpolated instant. To test this, we vary only the temporal context seen by
the trajectory estimator $g_\theta$ at inference. For each keyframe pair, we run a second
encoder pass restricted to $T$ keyframes centred on the pair, and feed these features to
$g_\theta$; the splines it predicts then drive the splatting step. All other features,
including the splatted ones, keep the full input window and are therefore identical across all
cases. To isolate the trajectory, we use the variant of our model without visibility
estimation (uniform splatting weights, Appendix~\ref{app:visibility}), trained under the same
setting as the full model. We compare against two closed-form references built on the same
optical flow: \emph{Linear Motion}, which scales the flow along straight lines, and
\emph{Quadratic Motion}, the constant-acceleration model of QVI~\citep{qvi}, which uses the
same four frames as $T{=}4$.
  
\begin{figure}[h]
  \centering
    \caption{PSNR $\uparrow$ on interpolated GoPro
  frames ($\times4$ spatial, $\times8$ temporal) as a function of the number of frames
  used to encode the features passed to the trajectory estimator. The splatted features always
  come from the full sequence.
  \emph{Linear Motion}: linear motion from optical flow. \emph{Quadratic Motion}: the 4-frame
  constant-acceleration model of QVI~\citep{qvi}. \emph{Full Window}: the unmodified model.}
  \includegraphics[width=0.48\textwidth]{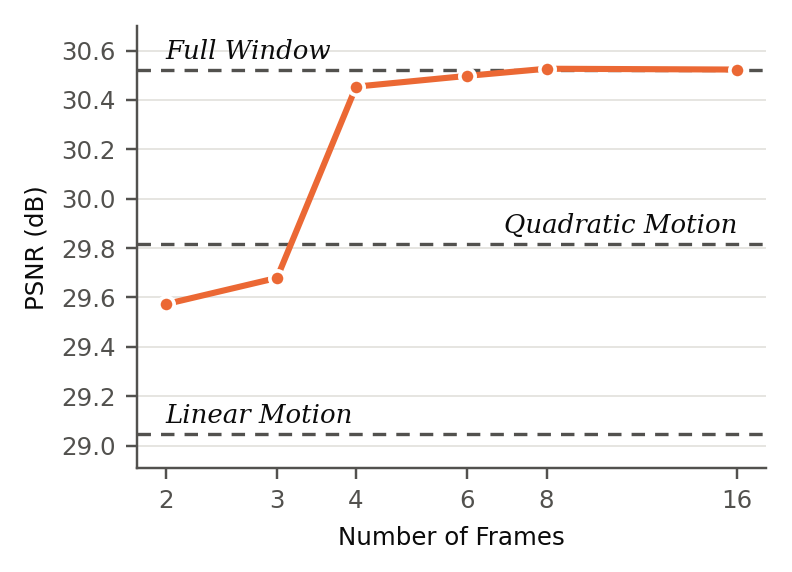}
  \label{fig:context-ablation}
\end{figure}

  Figure~\ref{fig:context-ablation} reports the results. With only a two-frame context, our learned trajectory already improves over linear motion on every clip. Performance then jumps once each keyframe of
  the pair sees a neighbour on its far side: at $T{=}4$, our trajectory recovers $96\%$ of
  the full-window gain over linear motion, against $52\%$ for QVI given the same frames,
  and outperforms it on every video. From $T{=}6$ on, results start to have diminishing returns. The trained estimator therefore relies on context beyond two frames.

Across all trajectory models, including linear and quadratic motion, LPIPS stays between
$0.1318$ and $0.1373$, while PSNR varies by up to $1.48$\,dB (Table~\ref{tab:context-lpips}):
replacing the trajectory at inference leaves perceptual quality essentially unchanged, and the
differences in PSNR reflect where content is placed rather than how it is rendered.

 \begin{table}[h]
  \caption{PSNR$\uparrow$ / LPIPS$\downarrow$ on interpolated GoPro frames ($\times4$, $\times8$)
  for each trajectory model of Figure~\ref{fig:context-ablation}. We mark in \textbf{bold} the best values and in \textit{italic} the worst.}
  \label{tab:context-lpips}
  \centering
  \begin{tabular}{l | c c}
  Trajectory & PSNR & LPIPS \\
  \hline
  Linear Motion    & \textit{29.05} & 0.1344 \\
  Quadratic Motion & 29.82 & \textit{0.1373} \\
  Ours, $T{=}2$    & 29.57 & 0.1342 \\
  Ours, $T{=}4$    & 30.45 & \textbf{0.1318} \\
  Ours, $T{=}6$    & 30.50 & 0.1327 \\
  Ours, $T{=}8$    & \textbf{30.53} & 0.1329 \\
  Ours, $T{=}16$   & 30.52 & 0.1329 \\
  Ours, Full Window      & 30.52 & 0.1329 \\
  \end{tabular}
  \end{table}

\section{Visibility Maps}
\label{app:visibility}
 When features from two input frames are warped to the same intermediate time, the frames can disagree about what is visible: an object may hide part of the background in one frame and reveal it in the other. Without visibility weights, the overlapping features are simply averaged. Because this averaging happens in feature space, the decoder can still resolve many of these conflicts, which already makes the model fairly robust to such conflicts. Visibility weights make the choice explicit, letting the model rely on whichever frame actually sees each region. Averaged over whole frames the improvement is small, partly because occlusions cover only a small part of each image, but it is consistent across all datasets (Table~\ref{tab:vis}). We therefore also evaluate on occlusion-boundary areas alone, selected where a forward–backward consistency check on RAFT flow fails. In these areas, the gain is about twice as large; this is where the mechanism is expected to act.
 
\ifdefined\vish\else\newlength\vish\fi
\begin{figure}[t]
  \centering
  \begin{minipage}[c]{0.34\linewidth}
    \centering
    {\footnotesize LR input frame $t$}\\[2pt]
    \includegraphics[width=\linewidth]{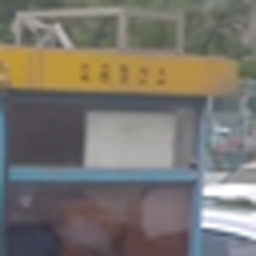}
  \end{minipage}\hfill
  \begin{minipage}[c]{0.64\linewidth}
    \centering
    \settoheight{\vish}{\includegraphics[width=0.31\linewidth]{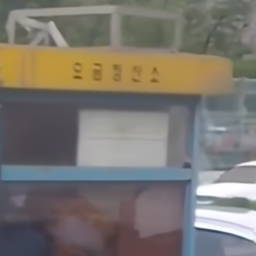}}
    \setlength{\tabcolsep}{1pt}
    \renewcommand{\arraystretch}{0}
    \begin{tabular}{@{}cccc@{}}
      & {\footnotesize $\tau{=}\tfrac14$} & {\footnotesize $\tau{=}\tfrac12$} & {\footnotesize $\tau{=}\tfrac34$} \\[3pt]
      \rotatebox{90}{\makebox[\vish][c]{\footnotesize Output}} &
      \includegraphics[width=0.31\linewidth,height=\vish]{figures/visibility/out_2.png} &
      \includegraphics[width=0.31\linewidth,height=\vish]{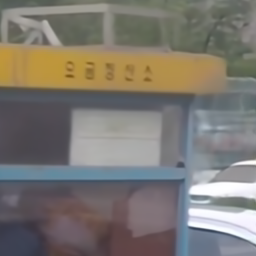} &
      \includegraphics[width=0.31\linewidth,height=\vish]{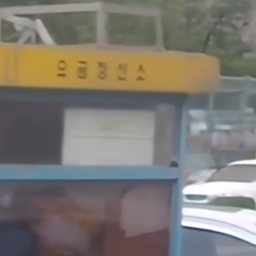} \\[2pt]
      \rotatebox{90}{\makebox[\vish][c]{\footnotesize Visibility}} &
      \includegraphics[width=0.31\linewidth,height=\vish]{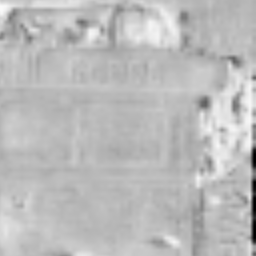} &
      \includegraphics[width=0.31\linewidth,height=\vish]{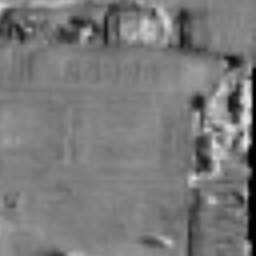} &
      \includegraphics[width=0.31\linewidth,height=\vish]{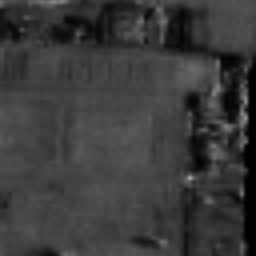} \\
    \end{tabular}
  \end{minipage}
  \caption{\textbf{Visibility over time.} Top: our output at three intermediate times between input
  frames $t$ and $t{+}1$. Bottom: the visibility weight of the features of frame $t$, splatted along
  their trajectories to the output time $t{+}\tau$, so it shows where frame $t$ contributes to each
  output (bright: high weight). Motion boundaries such as object edges are down-weighted the most. As $\tau$ grows, the weight decreases and frame $t$ contributes less.}
  \label{fig:vis}
\end{figure}

\begin{table}[h]\centering\small               
  \caption{\textbf{Effect of visibility weighting} (PSNR-Y, dB). \emph{Uniform}: overlapping                                                                  
  warped features are simply averaged. Both models share the same architecture and training setup. The last column scores only the occluded areas of GoPro.}                    \label{tab:vis}                               \begin{tabular}{@{}lcccc@{}}                                                                                                                                
  \toprule                                                                                                                                                    
   & GoPro & Adobe240 & Vid4 & GoPro (occluded areas) \\                                                                                                      
  \midrule                                                                                                                                                    
  Uniform    & 32.47          & 32.69          & 26.78          & 27.75 \\                                                                                    
  Visibility & \textbf{32.54} & \textbf{32.78} & \textbf{26.87} & \textbf{27.90} \\
  \bottomrule 
  \end{tabular} 
  \end{table}
\section{B-Spline curves}
\label{app:b-splines}

We use the cubic cardinal B-spline $\beta^{3}$, obtained by recursive
convolution of the box function $\beta^{0}(x) = \mathbf{1}_{[-\frac{1}{2}, \frac{1}{2})}(x)$ as
$\beta^{n} = \beta^{n-1} * \beta^{0}$:
\begin{equation}
\beta^{3}(x) =
\begin{cases}
\frac{2}{3} - x^{2} + \frac{|x|^{3}}{2}, & |x| < 1,\\[2pt]
\frac{(2-|x|)^{3}}{6}, & 1 \le |x| < 2,\\[2pt]
0, & |x| \ge 2.
\end{cases}
\end{equation}
It is non-negative, supported on $[-2,2]$ and $C^{2}$. A trajectory is built from $K$ atoms,
each translated by a knot $\xi_k$, dilated by $\lambda_k$ and weighted by a two-dimensional
control coefficient $c_k$ in $\mathbb{R}^{2}$ (one component per image axis, sharing the
knot). Any linear combination of such atoms is $C^{2}$ in time, so every trajectory has
continuous velocity and acceleration.

The uncentered spline is:
\begin{equation}
\rho(\tau;\lambda,\mu_t(u)) = \sum_{k=1}^{K} c_{t,k}(u)\,
\beta^{3}\!\big(\lambda_k\,(\tau-\xi_{t,k}(u))\big) \in \mathbb{R}^{2},
\end{equation}
and the residual used in the trajectories of Section~\ref{sec:trajectory} subtracts its value at
$\tau=0$, the time of the source frame $Y_t$:
\begin{equation}
s(\tau;\lambda,\mu_t(u)) = \rho(\tau;\lambda,\mu_t(u)) - \rho(0;\lambda,\mu_t(u)).
\end{equation}
Hence $s(0)=0$ and every trajectory starts exactly at $u_t$, whereas $s(1)$ is left free, so the
residual can correct the optical-flow endpoint $u_t + \vec{f}_t(u_t)$. Unlike scaling the residual
by $\tau$, centring pins the start point without damping the correction near it. The forward
residual is $\vec{s}(\tau, \lambda, \vec{\mu}_t(u)) = s(\tau;\lambda,\vec{\mu}_t(u))$. The backward residual
$\lvec{s}(\tau, \lambda,\lvec{\mu}_t(u)) = s(1-\tau;\lambda,\lvec{\mu}_t(u))$ is read at $1-\tau$, the time elapsed since
its source frame, so it vanishes at $\tau=1$, where the backward trajectory starts. The visibility weights of Section~\ref{sec:trajectory} follow the same convention: the backward weight $\lvec{w}_t$ is read at $1-\tau$.

\section{Additional Experiments}

\subsection{Arbitrary-Scale Video Super-Resolution}
\label{app:decoupled}
Following~\citep{v-cube}, we evaluate our method on the arbitrary-scale video
  super-resolution task (temporal factor $1$). We compare against C-STVSR
  methods~\citep{videoinr, motif, bfstvsr, v-cube}, the arbitrary-scale image SR methods
  LTE~\citep{lte}, CLIT~\citep{clit} and HIIF~\citep{hiif} applied per frame on an
  RDN~\citep{rdn} backbone, and the dedicated arbitrary-scale video SR method
  ST-AVSR~\citep{shangavsr}, on the 30 REDS~\citep{reds} validation videos ($100$ frames at
  $1280\times720$) with metrics computed on RGB following~\citep{shangavsr}.

Table~\ref{tab:reds-avsr} shows that our method is the best C-STVSR model at every scale, with the
  largest margins at $\times2$ and $\times3$. This holds even though these outputs are obtained by
  resampling our $\times4$ reconstruction (Section~\ref{sec:decoding}), which supports splatting on a
  fixed grid. Unlike ST-AVSR, which is trained on REDS~\citep{shangavsr}, all C-STVSR methods,
  including ours, are trained on Adobe240 and evaluated here across datasets. Despite this, our
  method surpasses ST-AVSR in PSNR at $\times2$ and $\times8$ and remains within $0.6$\,dB of it at
  the other scales. Results of competing methods are taken from~\citet{v-cube}.

\begin{table}[th]
\caption{Spatial video super-resolution (AVSR) on the REDS validation set (PSNR-RGB$\uparrow$ / SSIM-RGB$\uparrow$). Methods marked with $\dag$ indicate arbitrary-scale \textit{image} SR applied frame-by-frame for comparison; $\ddagger$ marks an arbitrary-scale \textit{video} SR method. We highlight the best result in \textbf{bold} and the second best with \underline{underline}.}
\label{tab:reds-avsr}
\centering
\begin{tabular}{l | c c c c c c}
Method & $\times 2$ & $\times 3$ & $\times 4$ & $\times 6$ & $\times 8$ & \# Par. \\
\hline \hline
Bicubic & 31.51 / 0.911 & 26.82 / 0.788 & 24.92 / 0.713 & 22.89 / 0.622 & 21.69 / 0.574 & --- \\
\hline
RDN-LTE $\dag$ & 34.73 / 0.943 & 30.73 / 0.866 & 28.75 / 0.804 & 26.56 / 0.718 & 25.24 / 0.669 & 22.5 M \\
RDN-CLIT $\dag$ & 34.63 / 0.942 & 30.63 / 0.865 & 28.63 / 0.801 & 26.43 / 0.714 & 25.14 / 0.661 & 37.7 M \\
RDN-HIIF $\dag$ & 34.57 / 0.942 & 30.59 / 0.864 & 28.60 / 0.799 & 26.44 / 0.712 & 25.15 / 0.659 & 23.2 M \\
ST-AVSR $\ddagger$& \underline{36.91} / \textbf{0.969} & \textbf{33.41}/\textbf{0.937} & \textbf{31.03} / \textbf{0.897} & \textbf{27.89} / \textbf{0.812} & 26.04 / \textbf{0.746} & 27.9 M\\
\hline
VideoINR & 31.59 / 0.900 & 30.04 / 0.852 & 28.13 / 0.791 & 25.27 / 0.687 & 23.46 / 0.619 & 11.3 M \\
MoTIF & 31.03 / 0.898 & 30.44 / 0.862 & 28.77 / 0.807 & 25.63 / 0.698 & 25.12 / 0.664 & 12.6 M \\
BF-STVSR & 34.72 / 0.946 & 31.17 / 0.881 & 29.11 / 0.820 & 26.78 / 0.728 & 25.40 / 0.668 & 13.5 M \\
V$^3$ & 36.53 / 0.963 & 32.31 / 0.908 & 29.92 / 0.849 & 27.39 / 0.754 & 25.96 / 0.690 & 13.7 M \\

V$^3$- Large & 36.70 / 0.964 & 32.53 / 0.911 & 30.13 / 0.853 & 27.54 / 0.760 & \underline{26.10} / 0.696 & 20.6 M \\

\rowcolor{lightgrayrow} 
Ours & \textbf{37.09} / \underline{0.967} &\underline{33.13} / \underline{0.921} & \underline{30.49} / \underline{0.863}
 & \underline{27.63} / \underline{0.764} & \textbf{26.11} / \underline{0.698} & 16.5 M
\end{tabular}
\end{table}

\subsection{Video Frame Interpolation}
\label{app:vfi}

We evaluate pure temporal upsampling on Adobe240 by feeding 30\,fps sequences at full spatial
resolution ($s=1$) and synthesizing 240\,fps output ($r=8$). Besides the C-STVSR methods,
we compare against two video frame interpolation (VFI) methods, RIFE~\citep{huang2022} and
EMA-VFI~\citep{ema-vfi}, using their official pretrained weights, trained on Vimeo-90K~\citep{vimeo90k} whereas all C-STVSR methods, ours included,
are trained on Adobe240. VFI methods return the input frames unchanged, while C-STVSR methods
reconstruct them, so the two families are compared on different columns of
Table~\ref{tab:vfi_x1t8}. Over all output frames, our method leads every C-STVSR method, by
2.4\,dB PSNR over V$^3$. On the seven synthesised frames between consecutive inputs, it
surpasses RIFE and EMA-VFI by 2.6 and 2.2\,dB PSNR.

\begin{table}[t]
\centering
  \caption{Temporal-only VSR with $\times 1$ spatial scale and $\times 8$ temporal scale on Adobe240. We report PSNR / SSIM on RGB outputs. ``Interp.'' averages only the 7 synthesised frames between consecutive inputs. $^\dagger$Frame-interpolation methods copy input frames unchanged, whereas C-STVSR methods reconstruct the keyframes.}                            
  \label{tab:vfi_x1t8}                                             
  \begin{tabular}{lcc}                                             
  \toprule                                                         
  Method & All frames & Interp. only \\ 
  \midrule                                                         
  VideoINR & 24.45 / 0.712 & -- \\                                 
  MoTIF    & 28.09 / 0.843 & -- \\                                 
  BF-STVSR & 29.37 / 0.867 & -- \\ 
  V$^3$    & \underline{33.43} / \underline{0.936} & -- \\  \midrule                                              RIFE$^\dagger$     & \textcolor{gray}{41.81 / 0.923} & 30.64 / 0.911 \\             
  EMA-VFI$^\dagger$  & \textcolor{gray}{42.16 / 0.929} & \underline{31.04} / \underline{0.918} \\            
  \midrule                                                        
  Ours     & \textbf{35.79} / \textbf{0.950} & \textbf{33.21} / \textbf{0.943} \\                  \bottomrule 
  \end{tabular}                                                      
  \end{table} 

\subsection{C-STVSR under degradations}
\label{app:degraded}

Following~\citet{v-cube}, we evaluate how C-STVSR methods trained without degradations, apart from
the space-time downscaling, handle noise and compression. Since the codec settings
of~\citet{v-cube} are not specified, we re-run the study for all methods under a common protocol.
Zero-mean Gaussian noise of standard deviation $\sigma$ is added to the input frames normalised to
$[-1,1]$, which are then clipped and quantised to 8 bits. In the compressed setting, the same noisy
frames are encoded as an H.264 stream (libx264, CRF~23, \texttt{medium} preset, YUV 4:2:0)
containing only the low-frame-rate input. Only the input is degraded; outputs are compared with the
clean ground truth.

All methods are fairly robust to noise (Table~\ref{tab:vid4-noise}), but not equally so.
The three splatting-based methods, MoTIF, BF-STVSR and ours, lose at most $0.17$\,dB up to
$\sigma{=}0.01$ and degrade by almost the same amount at every noise level, whereas V$^3$, which
decodes from the features of the nearest input frame without warping them, loses $0.5$\,dB at
$\sigma{=}0.01$. Since V$^3$ and our method share the same RVRT encoder, this points to the
reconstruction stage. Splatting forms each output feature as a weighted average of observations
from several frames, which share the scene content but carry independent noise, so the average
keeps the content and partly cancels the noise. As a result, our margin over V$^3$ grows from
$0.11$\,dB on clean inputs to $0.62$\,dB at $\sigma{=}0.025$. At $\sigma{=}0.05$, all methods
lose $1.3$--$1.6$\,dB.

Compression dominates the degradation (Table~\ref{tab:vid4-noise-h264}): all methods lose more
than $2$\,dB and end up on par. H.264 quantises the high-frequency transform coefficients most
coarsely, so at this rate much of the fine detail that multi-frame super-resolution relies on
is removed from the input, and blocking and ringing artefacts are added in its place. A model trained only on clean inputs has no way to tell these
artefacts apart from real content, so it reconstructs them rather than removing them.
Recovering detail from compressed video requires training on compressed inputs, as done by
dedicated compressed-VSR methods~\citep{comisr, realbasicvsr}; none of the evaluated methods,
ours included, is trained this way. The codec also explains why the noise level has almost no
further effect: quantisation removes most of the added noise along with the detail, so every
noise level produces nearly the same encoded input.

 \begin{table}[th]
  \caption{Robustness to Gaussian input noise on Vid4 ($\times4$ spatial, $\times2$ temporal). We mark the best result in \textbf{bold} and the second best \underline{underline}.}
  \label{tab:vid4-noise} 
  \centering
  \begin{tabular}{l | c | c c c c c}
   Method & Clean & $\sigma{=}0.005$ & $\sigma{=}0.0075$ & $\sigma{=}0.01$ & $\sigma{=}0.025$ & $\sigma{=}0.05$ \\
  \hline
  MoTIF    & 25.76 & 25.70 & 25.66 & 25.59    & 25.14    & 24.45    \\
  BF-STVSR & 25.84 & 25.80 & 25.75 & 25.69 & 25.26 & 24.52    \\
  V$^3$    & \underline{26.76} & \underline{26.61}    & \underline{26.42}    & \underline{26.24}    & \underline{25.69}    & \underline{25.21}    \\
  \rowcolor{lightgrayrow}
  \textbf{Ours} & \textbf{26.87} & \textbf{26.79} & \textbf{26.78} & \textbf{26.72} & \textbf{26.31} & \textbf{25.57} \\
  \end{tabular} 
  \end{table}

  \begin{table}[th]
  \caption{Robustness to noise followed by H.264 compression on Vid4 ($\times4$ spatial, $\times2$
temporal). All methods are on par.}
  \label{tab:vid4-noise-h264}
  \centering
  \begin{tabular}{l | c c c c c}
   Method & $\sigma{=}0.005$ & $\sigma{=}0.0075$ & $\sigma{=}0.01$ & $\sigma{=}0.025$ & $\sigma{=}0.05$ \\
  \hline
  MoTIF    & 23.56 & 23.55 & 23.56    & 23.53     & 23.47     \\
  BF-STVSR & 23.63 & 23.62 & 23.62 & 23.60 & 23.53 \\
  V$^3$    & 23.88    & 23.87    & 23.87    & 23.85    & 23.79    \\
  \rowcolor{lightgrayrow}
  \textbf{Ours} & 23.89 & 23.89 & 23.88 & 23.87 & 23.81 \\
  \end{tabular}
  \end{table}

\section{Additional metrics}
\label{app:additional-metrics}
\subsection{Runtime measurement}\label{app:runtime}
FPS is measured during the GoPro evaluation itself, on full-resolution test clips
($320{\times}180$ input, $1280{\times}720$ output) rather than crops, on a single RTX~6000 Ada
GPU. Each clip is processed in temporal windows, and only the model's forward pass is timed:
frame loading, host-to-device transfer, metric computation and output writing all fall
outside the timer. A warm-up before the first clip absorbs one-off costs such as compilation
and autotuning. Context frames recomputed at window boundaries are included in the model time (for V$^3$ and ours). FPS is the total number of output frames divided by the accumulated model
time over the whole test set.

V$^3$ reports a speed advantage over prior C-STVSR methods, measured on
$14{\times}80{\times}80$ crops. We could not reproduce this advantage on full frames: its
tiled encoder and per-patch decoder must then cover the entire $320{\times}180$ input, and
its throughput drops to 1.3\,FPS.
\subsection{Perceptual and temporal coherence metrics}

Following BF-STVSR~\citep{bfstvsr}, Table~\ref{tab:additionnal-metrics} reports four additional
video metrics, computed offline on the saved output frames with the same protocol.
VFIPS~\citep{vfips} and FloLPIPS~\citep{flolpips} are perceptual metrics designed for frame
interpolation: VFIPS is computed on non-overlapping 12-frame segments, and FloLPIPS on all
consecutive frame pairs after conversion to 8-bit YUV\,4:2:0. tOF~\citep{tof} measures temporal
consistency as the difference between the optical flows of the output and ground-truth videos.
VMAF~\citep{vmaf} is a video quality metric fitted to human opinion scores.

Our method is best on every metric except tOF on GoPro, where it is within $0.007$ of V$^3$                                                                 
  ($0.377$ vs.\ $0.370$); on Adobe240 it is also best in tOF (0.395 vs. 0.401). Temporal coherence                                                               
  is therefore on par with V$^3$, even though V$^3$ represents the video with space-time Fourier                                                              
  fields, which vary smoothly in time by construction, whereas we synthesise each frame by                                                                    
  splatting along predicted trajectories. The perceptual metrics favour our method more clearly:                                                              
  FloLPIPS is $11$--$12\%$ lower than V$^3$ on both datasets, consistent with the LPIPS gains of                                                              
  Table~\ref{tab:sf4tsf8-psnry-lpips}.

\begin{table}[htbp]
\centering
\caption{Comparison of one-stage C-STVSR methods on the GoPro and Adobe240 datasets. Results are evaluated using VFIPS~\citep{vfips} (negated so that higher is better), FloLPIPS~\citep{flolpips}, tOF~\citep{tof}, and VMAF~\citep{vmaf} metrics. All frames are upsampled ×4 spatially and ×8 temporally. In \textbf{bold} and \underline{underline} we respectively indicate the first and second best methods.}
\label{tab:additionnal-metrics}
\begin{tabular}{lcccccccc}
\toprule
\multicolumn{1}{c}{\raisebox{-1.5ex}{Method}} & \multicolumn{4}{c}{GoPro} & \multicolumn{4}{c}{Adobe} \\
\cmidrule(lr){2-5} \cmidrule(lr){6-9}
& VFIPS$\uparrow$ & FloLPIPS$\downarrow$ & tOF$\downarrow$ & VMAF$\uparrow$ & VFIPS$\uparrow$ & FloLPIPS$\downarrow$ & tOF$\downarrow$ & VMAF$\uparrow$ \\
\midrule
VideoINR & 81.13 & 0.151 & 0.519 & 57.96 & 81.15 & 0.145 & 0.574 & 67.08 \\
MoTIF & 81.89 & 0.156 & 0.517 & 59.82 & 81.61 & 0.144 & 0.607 & 68.40 \\
BF-STVSR & 83.01 & 0.151 & 0.480 & 61.06 & 84.04 & 0.132 & 0.498 & 70.82 \\
V$^3$ & \underline{88.21} & \underline{0.141} & \textbf{0.370} & \underline{70.88} & \underline{88.97} &\underline{0.113}& \underline{0.401}&\underline{78.40}\\
\rowcolor{lightgrayrow} 
Ours &\textbf{88.50}&\textbf{0.126}&\underline{0.377}&\textbf{71.76}& \textbf{90.24} & \textbf{0.099} & \textbf{0.395} & \textbf{79.40} \\
\bottomrule
\end{tabular}
\end{table}

\textbf{Video encoding for VMAF measurement.} Following the protocol from~\citet{bfstvsr}, we encode both the reconstructed and the ground-truth sequences to
H.264 with \texttt{libx264} at 30 frames per second in \texttt{yuv420p}, at ffmpeg's
default rate factor, and compute VMAF on the luma plane of the \emph{decoded}
streams. The codec is therefore part of the measurement for this metric: the
score reflects the reconstruction as it would be delivered, not the network
output in isolation.

\subsection{Luminance SSIM and RGB PSNR}

Table~\ref{tab:psnr-ssim} reports SSIM computed on luminance and PSNR computed on RGB. We recall that evaluating metrics on the luminance channel only is insensitive to artefacts such as colour bleeding, since the information needed to detect them lies in the chroma channels, which are ignored. 

\begin{table}[th]
  \caption{C-STVSR at $\times4$ spatial scale, $\times8$ temporal (GoPro, Adobe240) and $\times2$
temporal (Vid4). PSNR-RGB$\uparrow$ / SSIM-Y$\uparrow$. Methods marked $^\dagger$ have no public weights, so we report their published SSIM-Y only. We highlight the best result in \textbf{bold} and the second best with \underline{underline}.}                                                        
  \label{tab:psnr-ssim}                                                                                                                                       
  \centering                                                                                                             \setlength{\tabcolsep}{4pt}     
    \resizebox{\linewidth}{!}{%
      \begin{tabular}{l | l |l l l l l}
      \hline
      \multicolumn{1}{c}{Method} & \multicolumn{1}{c}{Params (M)} & \multicolumn{1}{c}{\parbox{2cm}{\centering {Vid4}\\}} &
      \multicolumn{1}{c}{\parbox{2cm}{\centering {GoPro}\\\it Center}} &
      \multicolumn{1}{c}{\parbox{2cm}{\centering {GoPro}\\\it Average}} & \multicolumn{1}{c}{\parbox{2cm}{\centering {Adobe}\\\it Center}} &
      \multicolumn{1}{c}{\parbox{2cm}{\centering {Adobe}\\\it Average}} \\
      \hline
      MoTIF & 12.6 & 24.23 / 0.774 & 29.52 / 0.888 & 28.53 / 0.877 & 29.07 / 0.884 & 28.27 / 0.875\\
      BF-STVSR & 13.5 & 24.35 / 0.777 & 29.67 / 0.890 & 28.74 / 0.880 & 29.31 / 0.888 & 28.60 / 0.880\\
      GS-STVSR$^\dagger$ & 12.7 & \phantom{0.}\text{---}\phantom{0} / \phantom{0.}\text{---}\phantom{0} & \phantom{0.}\text{---}\phantom{0} / 0.892 &
      \phantom{0.}\text{---}\phantom{0} / 0.882 & \phantom{0.}\text{---}\phantom{0} / 0.891 & \phantom{0.}\text{---}\phantom{0} / 0.883 \\
      V$^3$ & 13.7 & \underline{25.23} / \underline{0.819} & \underline{31.49} / 0.923 & \underline{30.77} / 0.919 & \underline{31.39} / 0.922 &
      \underline{30.73} / 0.917\\
      V$^3$- Large$^\dagger$ & 20.6 & \phantom{0.}\text{---}\phantom{0} / \phantom{0.}\text{---}\phantom{0} & \phantom{0.}\text{---}\phantom{0} /
  \underline{0.925} &
      \phantom{0.}\text{---}\phantom{0} / \underline{0.921} & \phantom{0.}\text{---}\phantom{0} / \underline{0.924} & \phantom{0.}\text{---}\phantom{0} /
      \underline{0.919}\\
      \textbf{Ours} & 16.5 & \textbf{25.34} / \textbf{0.826} & \textbf{31.95} / \textbf{0.929} & \textbf{31.07} / \textbf{0.924} & \textbf{32.06} /
      \textbf{0.931} & \textbf{31.26} / \textbf{0.925}\\
      \hline
      \end{tabular}%
    }                                                                                                                                                                     
  \end{table}

\section{Additional results}
\label{app:qualitative}

IMG\_0012 (Figure~\ref{fig:IMG_0012}) contains difficult motion: the camera pans non-uniformly while a large bus passes close to it. Our results look clearly sharper than those of V$^3$, but their PSNR is lower at $t{=}\tfrac12$:
15.3--20.0\,dB for ours against 17.8--21.3\,dB for V$^3$ over the three crops, whereas our LPIPS is about
40\% lower (0.185--0.243 against 0.304--0.396). The reason is
position, not appearance: our output is shifted by 4--6 pixels from the ground truth,
and PSNR strongly penalises a small shift of sharp edges, while a blurred output is penalised less.
To measure this, we search, for each crop and each method, for the integer translation within $\pm12$ pixels that
maximizes PSNR against the ground truth, and compute PSNR after applying it. Our PSNR then rises to
26.9--28.3\,dB, while V$^3$ only reaches 20.5--24.8\,dB, since shifting cannot recover blurred detail. On IMG\_0015 (Figure~\ref{fig:IMG_0015}) and IMG\_0004a (Figure~\ref{fig:IMG_0004a}), the best shift is zero, and
our method has the highest PSNR without any correction. 

The same pattern appears at the dataset level (Table~\ref{tab:sf4tsf8-psnry-lpips}): on GoPro and
Adobe240 at $\times8$, our PSNR-Y gain over V$^3$ is moderate (0.28--0.64\,dB), while our LPIPS is
12--17\% lower. This is consistent with sharp outputs that PSNR penalises for small displacements
but that a perceptual metric, which is more tolerant to such shifts, rewards.
\ifdefined\stcropw\else\newlength\stcropw\newlength\stcroph\newlength\stgroupw\fi
\begin{figure*}[t]
  \centering
  \includegraphics[width=0.5\linewidth]{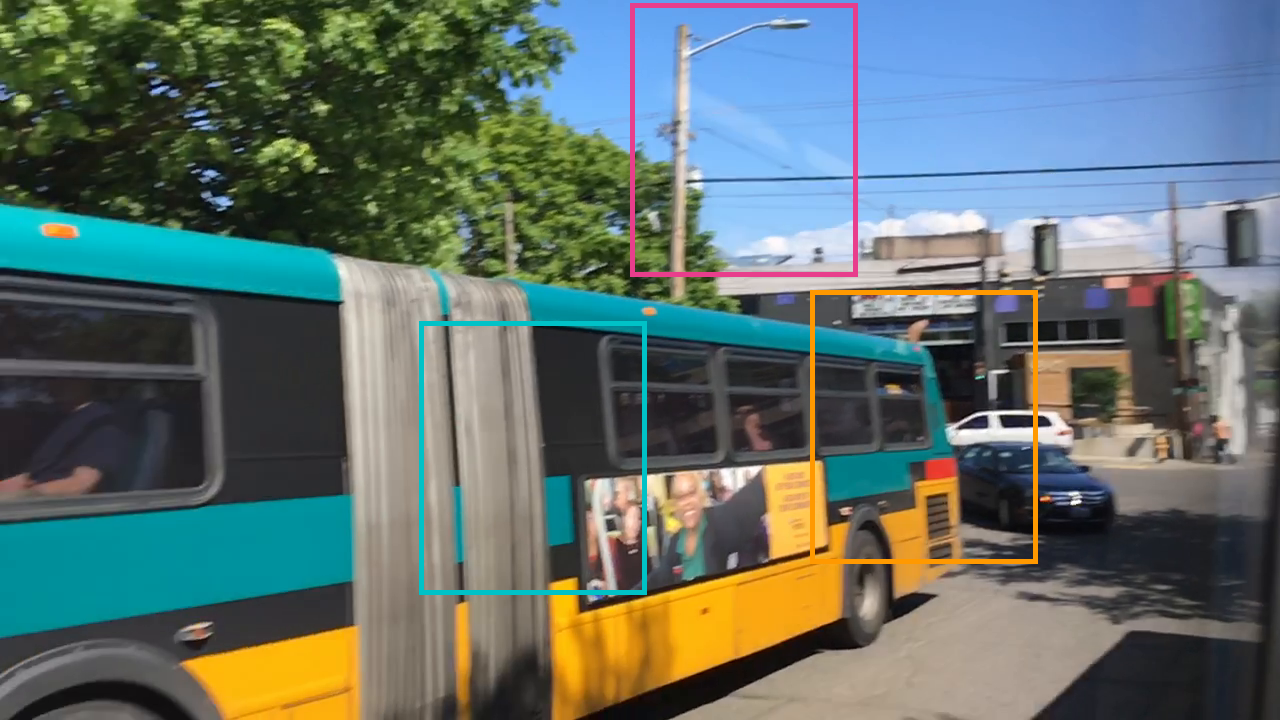}\par
  \vspace{3pt}
  {%
    \setlength{\fboxrule}{0.6pt}\setlength{\fboxsep}{0pt}%
    \setlength{\stcropw}{\dimexpr(\linewidth-1.1em-18\fboxrule-6\dimexpr1pt\relax-2\dimexpr4pt\relax-2pt)/9\relax}%
    \setlength{\stcroph}{\dimexpr\stcropw*273/227\relax}%
    \ifdim\stcroph>0.09\textheight\setlength{\stcroph}{0.09\textheight}\setlength{\stcropw}{\dimexpr\stcroph*227/273\relax}\fi
    \setlength{\stgroupw}{\dimexpr3\stcropw+6\fboxrule+2\dimexpr1pt\relax\relax}%
    \def\stcrop#1#2{\fcolorbox[HTML]{#1}{FFFFFF}{\includegraphics[width=\stcropw,height=\stcroph]{#2}}}%
    \def\stlabel#1{\makebox[1.1em][l]{\rotatebox{90}{\makebox[\stcroph][c]{\small #1}}}}%
    \setlength{\baselineskip}{0pt}\setlength{\lineskip}{1.5pt}\setlength{\lineskiplimit}{0pt}%
    \makebox[1.1em]{}\makebox[\stgroupw][c]{\small $\tau{=}\tfrac14$}\hspace{4pt}\makebox[\stgroupw][c]{\small $\tau{=}\tfrac12$}\hspace{4pt}\makebox[\stgroupw][c]{\small $\tau{=}\tfrac34$}\par
    \stlabel{MoTIF}\stcrop{FF9900}{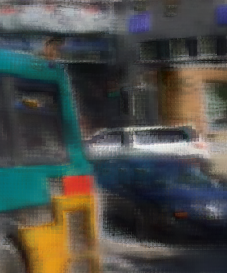}\hspace{1pt}\stcrop{00C2C7}{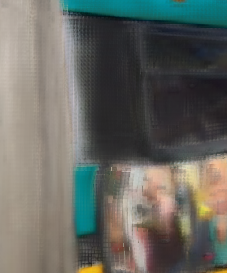}\hspace{1pt}\stcrop{E83E8C}{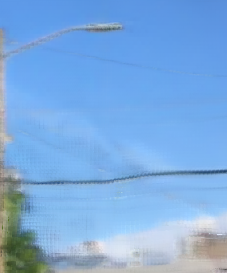}\hspace{4pt}\stcrop{FF9900}{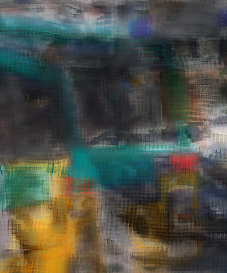}\hspace{1pt}\stcrop{00C2C7}{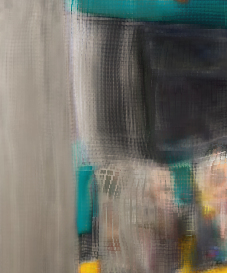}\hspace{1pt}\stcrop{E83E8C}{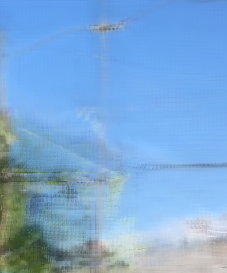}\hspace{4pt}\stcrop{FF9900}{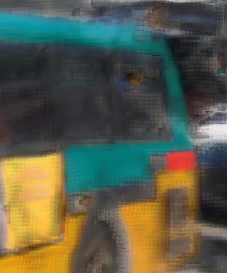}\hspace{1pt}\stcrop{00C2C7}{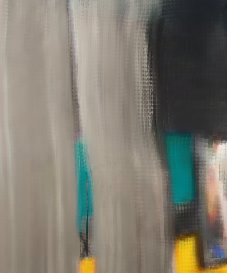}\hspace{1pt}\stcrop{E83E8C}{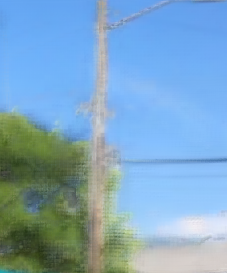}\par
    \stlabel{BF-STVSR}\stcrop{FF9900}{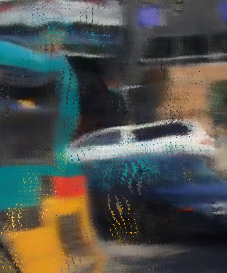}\hspace{1pt}\stcrop{00C2C7}{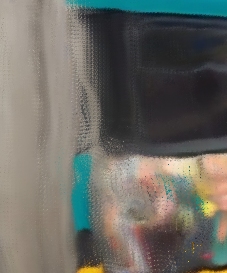}\hspace{1pt}\stcrop{E83E8C}{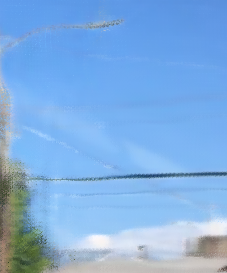}\hspace{4pt}\stcrop{FF9900}{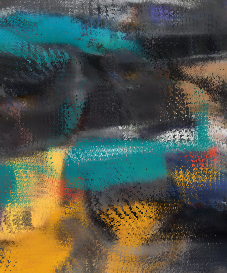}\hspace{1pt}\stcrop{00C2C7}{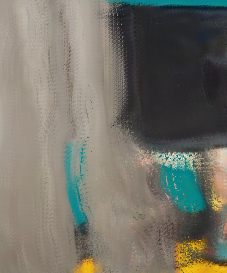}\hspace{1pt}\stcrop{E83E8C}{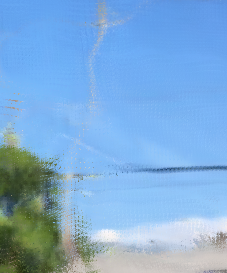}\hspace{4pt}\stcrop{FF9900}{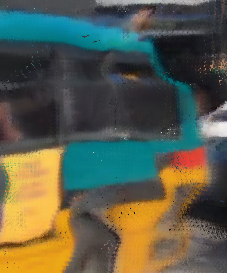}\hspace{1pt}\stcrop{00C2C7}{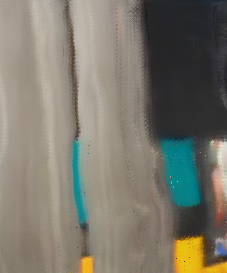}\hspace{1pt}\stcrop{E83E8C}{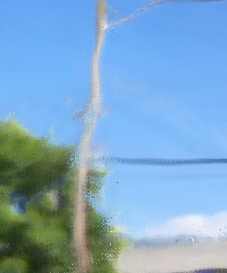}\par
    \stlabel{$V^3$}\stcrop{FF9900}{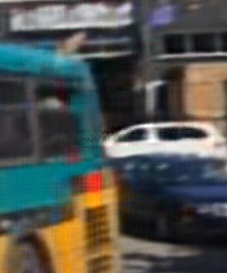}\hspace{1pt}\stcrop{00C2C7}{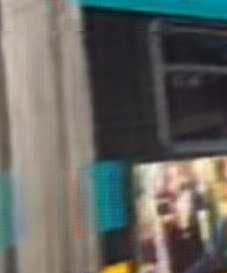}\hspace{1pt}\stcrop{E83E8C}{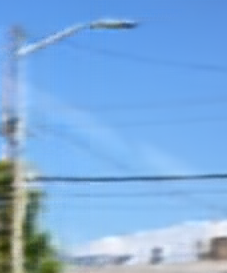}\hspace{4pt}\stcrop{FF9900}{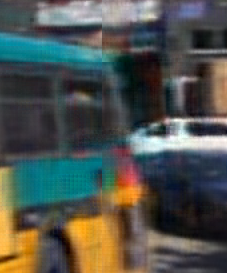}\hspace{1pt}\stcrop{00C2C7}{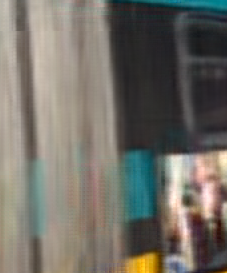}\hspace{1pt}\stcrop{E83E8C}{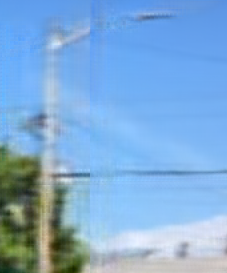}\hspace{4pt}\stcrop{FF9900}{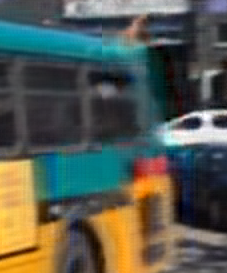}\hspace{1pt}\stcrop{00C2C7}{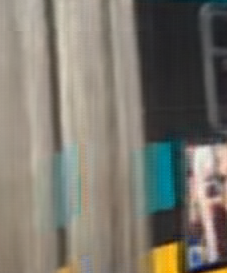}\hspace{1pt}\stcrop{E83E8C}{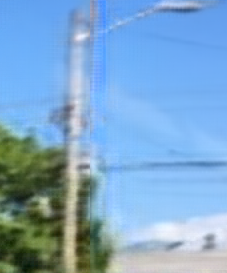}\par
    \stlabel{Ours}\stcrop{FF9900}{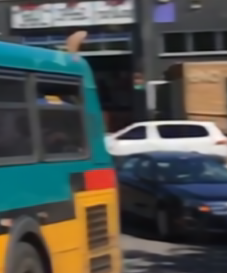}\hspace{1pt}\stcrop{00C2C7}{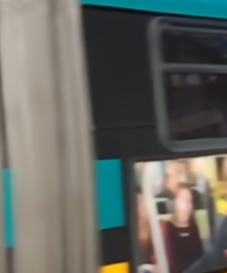}\hspace{1pt}\stcrop{E83E8C}{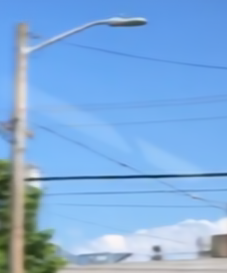}\hspace{4pt}\stcrop{FF9900}{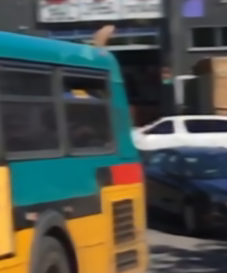}\hspace{1pt}\stcrop{00C2C7}{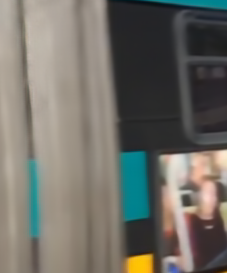}\hspace{1pt}\stcrop{E83E8C}{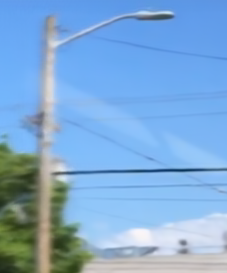}\hspace{4pt}\stcrop{FF9900}{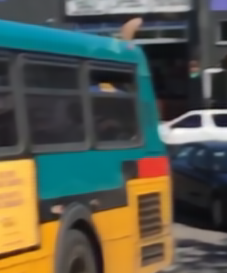}\hspace{1pt}\stcrop{00C2C7}{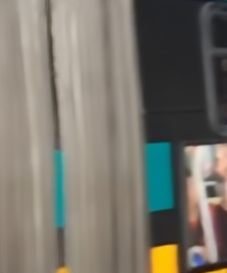}\hspace{1pt}\stcrop{E83E8C}{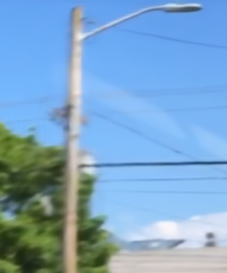}\par
    \stlabel{GT}\stcrop{FF9900}{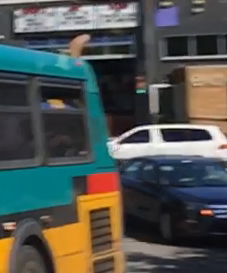}\hspace{1pt}\stcrop{00C2C7}{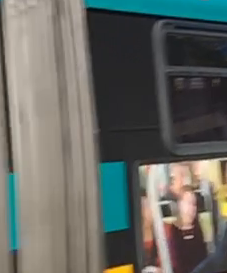}\hspace{1pt}\stcrop{E83E8C}{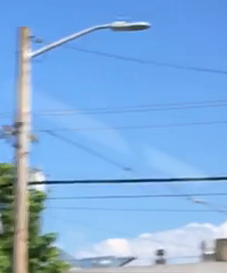}\hspace{4pt}\stcrop{FF9900}{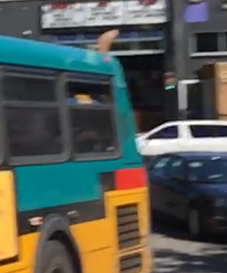}\hspace{1pt}\stcrop{00C2C7}{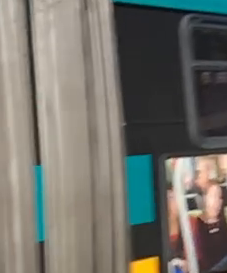}\hspace{1pt}\stcrop{E83E8C}{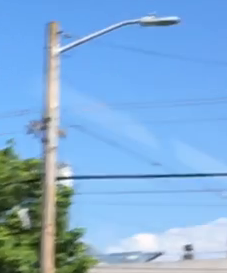}\hspace{4pt}\stcrop{FF9900}{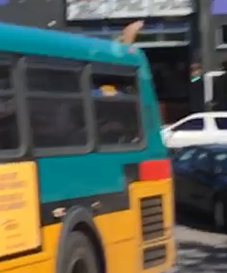}\hspace{1pt}\stcrop{00C2C7}{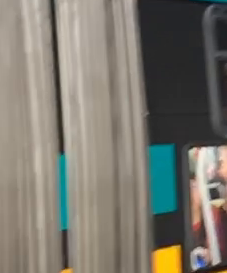}\hspace{1pt}\stcrop{E83E8C}{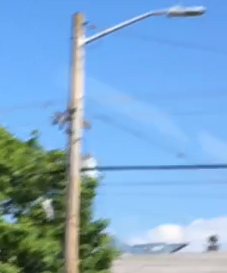}\par
  }
  \caption{Qualitative comparison on Adobe240 IMG\_0012 ($\times$4 space, $\times$8 time), at the
intermediate times between input keyframes 1721 and 1729. MoTIF and BF-STVSR break the bus front into
fragmented patches and $V^3$ blurs it with a visible seam; ours preserves the window frames and the back of the bus. Thin structures are the hardest case: the baselines lose the streetlight and most of
the power lines (MoTIF and BF-STVSR drop the pole entirely at $\tau{=}\tfrac14$), whereas ours keeps the pole,
the lamp arm and the individual wires at every intermediate time. Recovered details (or lack thereof) may be better appreciated by zooming on the figure.}
  \label{fig:IMG_0012}
\end{figure*}

\end{document}